\documentclass{aamas2011_cameraReady}
\usepackage{bm}
\newtheorem{theorem}{Theorem}
\newtheorem{lemma}[theorem]{Lemma}

\newdef{definition}{Definition}

\begin{document}
\newcommand{\ma}[1]{{\bf #1}}
\newcommand{\ve}[1]{{\mathbf #1}}
\newcommand{\set}[1]{{\mathcal #1}}
\newcommand{\defeq}[0]{\ensuremath{\stackrel{\triangle}{=}}}

\newcommand{\squishlist}{
 \begin{list}{$\bullet$}
  { \setlength{\itemsep}{5pt}
     \setlength{\parsep}{0pt}
     \setlength{\topsep}{0pt}
     \setlength{\partopsep}{0pt}
     \setlength{\leftmargin}{0.7em}
     \setlength{\labelwidth}{0.5em}
     \setlength{\labelsep}{0.2em} } }

\newcommand{\squishlisttwo}{
 \begin{list}{$\bullet$}
  { \setlength{\itemsep}{0pt}
     \setlength{\parsep}{0pt}
    \setlength{\topsep}{0pt}
    \setlength{\partopsep}{0pt}
    \setlength{\leftmargin}{1em}
    \setlength{\labelwidth}{1.5em}
    \setlength{\labelsep}{0.5em} } }

\newcommand{\squishend}{
  \end{list}  }



\AuthorsForCitationInfo{Kian Hsiang Low, John M. Dolan, and Pradeep Khosla}

\TitleForCitationInfo{Active Markov Information-Theoretic Path Planning for Robotic Environmental Sensing}

\title{Active Markov Information-Theoretic Path Planning for Robotic Environmental Sensing}



%
%
%
%

%

\numberofauthors{2}

\author{
%
\alignauthor
Kian Hsiang Low\\
       \affaddr{Department of Computer Science}\\
       \affaddr{National University of Singapore}\\
       \affaddr{Republic of Singapore}\\
       \email{lowkh@comp.nus.edu.sg}
\alignauthor
John M. Dolan and Pradeep Khosla\\
       \affaddr{Robotics Institute}\\
       \affaddr{Carnegie Mellon University}\\       
       \affaddr{Pittsburgh PA 15213 USA}\\
       \email{jmd@cs.cmu.edu, pkk@ece.cmu.edu}
}

\maketitle

\begin{abstract}
Recent research in multi-robot exploration and mapping has focused on sampling environmental fields, which are typically modeled using the Gaussian process (GP). Existing information-theoretic exploration strategies for learning GP-based environmental field maps adopt the non-Markovian problem structure and consequently scale poorly with the length of history of observations.
Hence, it becomes computationally impractical to use these strategies for \emph{in situ}, real-time active sampling.
To ease this computational burden, this paper presents a Markov-based approach to efficient information-theoretic path planning for active sampling of GP-based fields.
We analyze the time complexity of solving the Markov-based path planning problem, and demonstrate analytically that it scales better than that of deriving the non-Markovian strategies with increasing length of planning horizon.
For a class of exploration tasks called the transect sampling task, we provide theoretical guarantees on the active sampling performance of our Markov-based policy, from which ideal environmental field conditions and sampling task settings can be established to limit its performance degradation due to violation of the Markov assumption.
Empirical evaluation on real-world temperature and plankton density field data shows that our Markov-based policy can generally achieve active sampling performance comparable to that of the widely-used non-Markovian greedy policies under less favorable realistic field conditions and task settings while enjoying significant computational gain over them.\end{abstract}


\vspace{-1mm}



\terms{\vspace{-1mm}Algorithms, Performance, Experimentation, Theory}\vspace{-1mm}


\keywords{\vspace{-1mm}Multi-robot exploration and mapping, adaptive sampling, active learning, Gaussian process, non-myopic path planning}%

\section{Introduction}
\label{sect:intro}
Research in multi-robot exploration and mapping has recently progressed from building occupancy grids \cite{Thrun05} to sampling spatially varying environmental phenomena \cite{LowAAMAS08,LowICAPS09}, in particular,
environmental fields (e.g., plankton density, pollutant concentration, temperature fields) that are characterized by \emph{continuous-valued, spatially correlated} measurements (see Fig.~$1$).
Exploration strategies for building occupancy grid maps usually operate under the assumptions of (a) \emph{discrete}, (b) \emph{independent} cell occupancies, which impose, respectively, the following limitations for learning environmental field maps:
these strategies (a) cannot be fully informed by the continuous field measurements and (b) cannot exploit the spatial correlation structure of an environmental field for selecting observation paths.
As a result, occupancy grid mapping strategies are not capable of 
selecting the most informative observation paths for learning an environmental field map.

Furthermore, occupancy grid mapping strategies typically assume that range sensing is available.
In contrast, many \emph{in situ} environmental and ecological sensing applications (e.g., monitoring of ocean phenomena, forest ecosystems, or pollution) permit only point-based sensing, thus making a high-resolution sampling of the entire field impractical in terms of resource costs (e.g., energy consumption, mission time).
In practice, the resource cost constraints restrict the spatial coverage of the observation paths.
Fortunately, the spatial correlation structure of an environmental field enables 
a map of the field (in particular, its unobserved areas) to be learned
using the point-based observations taken along the resource-constrained paths.
To learn this map, a commonly-used approach in spatial statistics \cite{Webster01} is to assume that the environmental field is realized from a probabilistic model called the \emph{Gaussian process} (GP) (Section~\ref{sect:gp}).
More importantly, the GP model allows an environmental field to be formally characterized and consequently provides formal measures of mapping uncertainty (e.g., based on mean-squared error \cite{LowAAMAS08} or entropy criterion \cite{LowICAPS09}) for directing a robot team to explore highly uncertain areas of the field.
In this paper, we focus on using the entropy criterion to measure mapping uncertainty.

How then does a robot team plan the most informative resource-constrained observation paths to minimize the mapping uncertainty of an environmental field?
To address this, the work of \cite{LowICAPS09} has proposed an information-theoretic multi-robot exploration strategy
that selects non-myopic observation paths with maximum entropy.
Interestingly, this work has established an equivalence result that the maximum-entropy paths selected by such a strategy can achieve the dual objective of minimizing the mapping uncertainty defined using the entropy criterion.
When this strategy is applied to sampling a GP-based environmental field, it can be reduced to solving a non-Markovian, deterministic planning problem called the \emph{information-theoretic multi-robot adaptive sampling problem} ($i$MASP) (Section~\ref{sect:nmpp}).
Due to the non-Markovian problem structure of $i$MASP, its state size grows exponentially with the length of planning horizon.
To alleviate this computational difficulty, an anytime heuristic search algorithm called Learning Real-Time A$^{\ast}$ \cite{Korf90} is used to solve $i$MASP approximately. 
However, this algorithm does not guarantee the performance of its induced exploration policy. 
We have also observed through experiments
that when the joint action space of the robot team is large or the planning horizon is long,
it no longer produces a good policy fast enough. 
Even after incurring a huge amount of time and space to improve the search, its resulting policy still performs worse than the widely-used non-Markovian greedy policy, the latter of which can be derived efficiently by solving the myopic formulation of $i$MASP (Section~\ref{sect:imasp}).

Though the anytime and greedy algorithms provide some computational relief to solving $i$MASP (albeit approximately), they inherit $i$MASP's non-Markovian problem structure and consequently 
scale poorly with the length of history of observations.
Hence, it becomes computationally impractical to use these non-Markovian path planning algorithms for \emph{in situ}, real-time active sampling performed (a) at high resolution (e.g., due to high sensor sampling rate or large sampling region), (b) over dynamic features of interest (e.g., algal blooms, oil spills), 
(c) with resource cost constraints (e.g., energy consumption, mission time), or (d) in the presence of dynamically changing external forces translating the robots (e.g., ocean drift on autonomous boats), thus requiring fast replanning.
For example, the deployment of autonomous underwater vehicles (AUVs) and boats for ocean sampling poses the above challenges/issues among others \cite{Leonard07}.

To ease this computational burden, this paper proposes a principled Markov-based approach to efficient information-theoretic path planning for active sampling of GP-based environmental fields (Section~\ref{sect:mpp}), which we develop by assuming the Markov property in $i$MASP planning.
To the probabilistic robotics community, such a move to achieve time efficiency is probably anticipated.
However, the Markov property is often imposed without carefully considering or formally analyzing its consequence on the performance degradation while operating in non-Markovian environments.
In particular, to what extent does the environmental structure affect the performance degradation due to violation of the Markov assumption?
Motivated by this lack of treatment, our work in this paper is novel
in demonstrating both theoretically and empirically the extent of which the degradation of active sampling performance depends on the spatial correlation structure of an environmental field.
An important practical consequence is that of establishing environmental field conditions under which the Markov-based approach performs well relative to the non-Markovian $i$MASP-based policy while enjoying significant computational gain over it. 
The specific contributions of our work include:

\squishlisttwo
\item analyzing the time complexity of solving the Markov-based information-theoretic path planning problem, and showing analytically that it scales better than that of deriving the non-Markovian strategies with increasing length of planning horizon
(Section~\ref{sect:time});
\item providing theoretical guarantees on the active sampling performance of our Markov-based policy (Section~\ref{sect:guarantee})
for a class of exploration tasks called the \emph{transect sampling task} (Section~\ref{sect:tstask}),
from which various ideal environmental field conditions and sampling task settings can be established to limit its performance degradation;
\item empirically evaluating the active sampling performance and time efficiency of our Markov-based policy on real-world temperature and plankton density field data
under less favorable realistic environmental field conditions and sampling task settings  (Section~\ref{sect:expt}).
\squishend
\section{Transect Sampling Task}
\label{sect:tstask}
Fig.~\ref{fig:tempgrid} illustrates the transect sampling task introduced in \cite{Lamas00,Thompson08} previously. A temperature field is spatially distributed over a 25~m~$\times$~150~m transect that is discretized into a $5\times 30$ grid of sampling locations comprising $30$ columns, each of which has $5$ sampling locations.
It can be observed that the number of columns is much greater than the number of sampling locations in each column; this observed property is assumed to be consistent with every other transect.
The robots are constrained to simultaneously explore forward one column at a time from the leftmost to the rightmost column of the transect such that each robot samples one location per column for a total of $30$ locations.
So, each robot's action space given its current location consists of moving to any of the $5$ locations in the adjacent column on its right.
The number of robots is assumed not to be larger than the number of sampling locations per column.
We assume that an adversary chooses the starting robot locations in the leftmost column and the robots will only know them at the time of deployment; such an adversary can be the dynamically changing external forces translating the robots (e.g., ocean drift on autonomous boats) or the unknown obstacles occupying potential starting locations.
The robots are allowed to end at any location in the rightmost column.

In practice, the constraint on exploring forward in a transect sampling task permits the planning of less complex observation paths that can be achieved more reliably, using less sophisticated control algorithms, and by robots with limited maneuverability (e.g., unmanned aerial vehicles, autonomous boats and AUVs \cite{Davis04}).
For practical applications, while the robot is in transit from its current location to a distant planned waypoint \cite{Leonard07,Thompson08}, this task can be performed to collect the most informative observations during transit.
In monitoring of ocean phenomena and freshwater quality along rivers, the transect can span a plankton density or temperature field drifting at a constant rate from right to left and the autonomous boats are tasked to explore within a line perpendicular to the drift.
As another example, the transect can be the bottom surface of ship hull or other maritime structure to be inspected and mapped by AUVs.
%
\section{Non-Markovian Path Planning}
\label{sect:nmpp}
\subsection{Notations and Preliminaries} 
\begin{figure}
\centering
\hspace{-0mm}\epsfig{figure=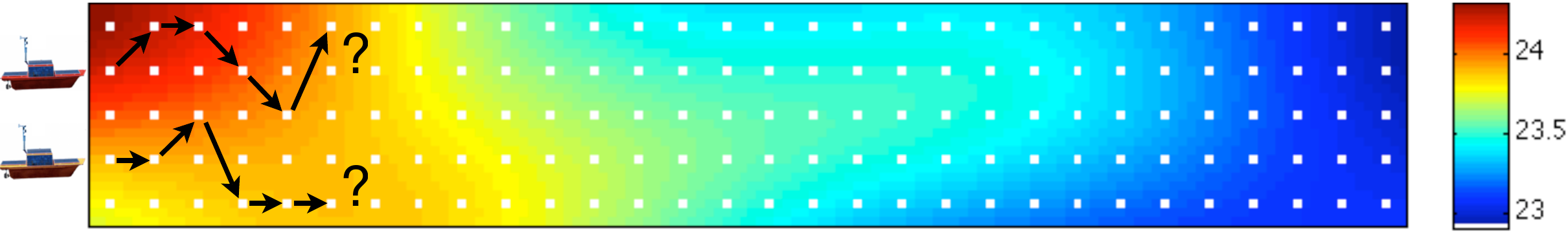,width=8.5cm}\vspace{-4mm}\\
\caption{Transect sampling task on a temperature field (measured in$\,^{\circ}\mathrm{C}$) spatially distributed over a $25$~m~$\times$~$150$~m transect that is discretized into a $5\times 30$ grid of sampling locations (white dots).
\vspace{-4mm}
}
\label{fig:tempgrid}
\end{figure}
Let $\set{U}$ be the domain of the environmental field representing a set of sampling locations in the transect such that each location $u\in\set{U}$ yields a measurement $z_u$.
The columns of the transect are indexed in an increasing order from left to right with the leftmost column being indexed `$0$'.
Each planning stage is associated with a column from which every robot in the team selects and takes an observation (i.e., comprising a pair of location and its measurement).
Let $k$ denote the number of robots in the team.
In each stage $i$, the team of $k$ robots then collects from column $i$ a total of $k$ observations, which are denoted by a pair of vectors $x_i$ of $k$ locations 
and $z_{x_i}$ of the corresponding measurements.
Let $x_{0:i}$ and $z_{x_{0:i}}$ denote vectors comprising the histories of robots' sampling locations and corresponding measurements over stages $0$ to $i$
 (i.e., concatenations of $x_0, x_1, \ldots, x_i$ and $z_{x_0}, z_{x_1}, \ldots, z_{x_i}$), respectively.
Let $Z_u$, $Z_{x_i}$, and $Z_{x_{0:i}}$ be random measurements that are associated with the realizations $z_u$, $z_{x_i}$, and $z_{x_{0:i}}$, respectively.
\vspace{-0mm}
\subsection{Gaussian Process-Based Environmental Field}
\label{sect:gp}
The GP model can be used to formally characterize an environmental field as follows:
the environmental field is defined to vary as a realization of a GP.
Let $\{Z_u\}_{u \in \set{U}}$ denote a GP, i.e., every finite subset of $\{Z_u\}_{u \in \set{U}}$ has a multivariate Gaussian distribution \cite{Rasmussen06}.
The GP is fully specified by its mean $\mu_{u} \defeq \mathbb{E}[Z_u]$ and covariance
$\sigma_{u v} \defeq \mbox{cov}[Z_u, Z_{v}]$
for all $u, v \in \set{U}$.
We assume that the GP is second-order stationary, i.e., it has a constant mean and a stationary covariance structure
(i.e., $\sigma_{u v}$ is a function of $u - v$ for all $u, v \in \set{U}$).
In particular, its covariance structure is defined by the widely-used squared exponential covariance function \cite{Rasmussen06}
\vspace{-1mm}
\begin{equation}
\sigma_{u v} \defeq \displaystyle \sigma^2_s \exp\left\{-\frac{1}{2}
(u-v)^\top M^{-2} (u-v)
\right\} + \sigma^2_n \delta_{uv}
\vspace{-1mm}
\label{eq:sexp3}
\end{equation}
where $\sigma^2_s$ is the signal variance, $\sigma^2_n$ is the noise variance, $M$ is a diagonal matrix with 
length-scale components $\ell_1$ and $\ell_2$ in the horizontal and vertical directions of a transect, respectively, and $\delta_{uv}$ is a Kronecker delta of value $1$ if $u = v$, and $0$ otherwise.
Intuitively, the signal and noise variances describe, respectively, the intensity and noise of the field measurements while the length-scale can be interpreted as the approximate distance to be traversed in a transect for the field measurement to change considerably \cite{Rasmussen06}; it therefore controls the degree of spatial correlation or ``similarity'' between field measurements.
In this paper, the mean and covariance structure of the GP are assumed to be known. 
Given that the robot team has collected observations $x_0, z_{x_0}, x_1, z_{x_1}, \ldots, x_i, z_{x_i}$ over stages $0$ to $i$, the distribution of $Z_u$ remains Gaussian with the following posterior mean and covariance
\vspace{-1mm}
\begin{equation}
\displaystyle 
\mu_{u \mid x_{0:i}}
= \mu_{u} + \Sigma_{u x_{0:i}} \Sigma^{-1}_{x_{0:i} x_{0:i}} \{ z_{x_{0:i}} - {\mu}_{{x_{0:i}}} \}^{\top}
\vspace{-2mm}
\label{eq:6}
\end{equation}
\begin{equation}
\displaystyle \sigma_{u v \mid x_{0:i}} = \sigma_{u v} - \Sigma_{u x_{0:i}} \Sigma^{-1}_{x_{0:i} x_{0:i}} \Sigma_{x_{0:i} v}
\vspace{-1mm}
\label{eq:7}
\end{equation}
where ${\mu}_{{x_{0:i}}}$
is a row vector with mean components $\mu_{{w}}$ for every location $w$ of $x_{0:i}$,
$\Sigma_{u {x}_{0:i}}$ is a row vector with covariance components $\sigma_{u w}$ for every location $w$ of $x_{0:i}$,
$\Sigma_{{x}_{0:i} v}$ is a column vector with covariance components $\sigma_{w v}$ for every location $w$ of $x_{0:i}$,
and $\Sigma_{x_{0:i} x_{0:i}}$ is a covariance matrix with components $\sigma_{w y}$ for every pair of locations $w, y$ of $x_{0:i}$.
Note that the posterior mean $\mu_{u \mid x_{0:i}}$ (\ref{eq:6}) is the best unbiased predictor of the measurement $z_{u}$ at unobserved location $u$. 
An important property of GP is that the posterior covariance $\sigma_{u v \mid x_{0:i}}$ (\ref{eq:7}) is independent of the observed measurements $z_{{x}_{0:i}}$; this property is used to reduce $i$MASP to a deterministic planning problem as shown later.
\subsection{Deterministic {\large{$i$}}MASP Planning}
\label{sect:imasp}
Supposing the robot team starts in locations $x_0$ of leftmost column $0$, 
an exploration policy is responsible for directing it to sample locations $x_1, x_2,\ldots,x_{t+1}$ of the respective columns $1, 2, \ldots, t+1$
to form the observation paths.
Formally, a non-Markovian policy is denoted by
$\pi \defeq\langle\pi_0(x_{0:0}=x_0),\pi_1(x_{0:1}),\ldots, \pi_{t}(x_{0:t})\rangle$ where $\pi_i(x_{0:i})$ maps the history $x_{0:i}$ of robots' sampling locations to a vector ${a}_i \in \set{A}({x}_i)$ of robots' actions in stage $i$ (i.e., ${a}_i\leftarrow\pi_i(x_{0:i})$), and
$\set{A}({x}_i)$ is the joint action space of the robots given their current locations ${x}_i$. We assume that the transition function $\tau({x}_i, {a}_i)$ \emph{deterministically} (i.e., no localization uncertainty) moves the robots 
to their next locations ${x}_{i+1}$ in stage $i+1$ (i.e., ${x}_{i+1}\leftarrow\tau({x}_i, {a}_i)$).
Putting $\pi_i$ and $\tau$ together yields the assignment ${x}_{i+1} \leftarrow \tau({x}_{i}, \pi_i(x_{0:i}))$.

The work of \cite{LowICAPS09} has proposed a non-Markovian policy $\pi^\ast$ that selects non-myopic observation paths with maximum entropy for sampling a GP-based field. To know how $\pi^\ast$ is derived, we first define the value under a policy $\pi$ to be the entropy of observation paths when starting in $x_0$ and following $\pi$ thereafter:
\begin{equation}
\hspace{-0mm}
\begin{array}{rl}
\displaystyle V^{\pi}_{0}({x}_{0}) \defeq& \hspace{-0mm}\displaystyle
\mathbb{H}[{Z}_{{x}_{1:t+1}}| Z_{{x}_{0}}, \pi]\\
=& \hspace{-0mm}- \displaystyle \int f(z_{x_{0:t+1}}|\pi) \log f(z_{x_{1:t+1}}|z_{x_{0}}, \pi) \ d z_{x_{0:t+1}}
\end{array}
\label{eq:9.9}
\end{equation}
where $f$ denotes a Gaussian probability density function.
When a non-Markovian policy $\pi$ is plugged into (\ref{eq:9.9}), the following $(t+1)$-stage recursive formulation results from the chain rule for entropy and ${x}_{i+1} \leftarrow \tau({x}_{i}, \pi_i(x_{0:i}))$:
\begin{equation}
\hspace{-0mm}
\begin{array}{rl}
\displaystyle V^{\pi}_{i}({x}_{0:i}) =& \hspace{-0mm}\displaystyle
\mathbb{H}[{Z}_{{x}_{i+1}}| Z_{{x}_{0:i}}, \pi_i] + V^{\pi}_{i+1}({x}_{0:i+1})\vspace{1mm}
\\
=& \hspace{-0mm}\displaystyle
\mathbb{H}[{Z}_{\tau({x}_{i}, \pi_i({x}_{0:i}))}| Z_{{x}_{0:i}}] +V^{\pi}_{i+1}( ({x}_{0:i}, \tau({x}_{i}, \pi_i({x}_{0:i}))) )
\vspace{1mm}
\\
\displaystyle V^{\pi}_{t}({x}_{0:t}) =& \hspace{-0mm}\displaystyle
\mathbb{H}[{Z}_{{x}_{t+1}}| Z_{{x}_{0:t}}, \pi_t]
\vspace{1mm}
\\ 
=& \hspace{-0mm}\mathbb{H}[{Z}_{\tau({x}_{t}, \pi_t({x}_{0:t}))}| Z_{{x}_{0:t}}]
\end{array}
\label{eq:9.9a}
\end{equation}
for stage $i = 0,\ldots,t-1$
such that each stagewise posterior entropy (i.e., of the measurements $Z_{x_{i+1}}$ to be observed in stage $i+1$ given the history of measurements $Z_{x_{0:i}}$ observed from stages $0$ to $i$) reduces to
\begin{equation}
\mathbb{H}[Z_{x_{i+1}}| Z_{x_{0:i}}] = \displaystyle\frac{1}{2}\log \ (2\pi e)^k |\Sigma_{x_{i+1}\mid x_{0:i}}|
\label{eq:7a}
\end{equation}
where $\Sigma_{x_{i+1}\mid x_{0:i}}$ is a covariance matrix with components $\sigma_{u v\mid x_{0:i}}$ for every pair of locations $u, v$ of $x_{i+1}$, each of which is independent of observed measurements $z_{{x}_{0:i}}$ by (\ref{eq:7}), as discussed above. So, $\mathbb{H}[Z_{x_{i+1}}| Z_{x_{0:i}}]$ can be evaluated in closed form, and the value functions (\ref{eq:9.9a}) only require the history of robots' sampling locations $x_{0:i}$ as inputs but not that of corresponding measurements $z_{{x}_{0:i}}$.
 
Solving $i$MASP involves choosing $\pi$ to maximize $V^{\pi}_{0}(x_0)$ (\ref{eq:9.9a}), which yields the optimal policy $\pi^\ast$. Plugging $\pi^\ast$ into (\ref{eq:9.9a}) gives the $(t+1)$-stage dynamic programming equations:
\begin{equation}
\begin{array}{rl}
\displaystyle V^{\pi^{\ast}}_{i}({x}_{0:i}) =& \hspace{-0mm}\displaystyle\max_{{a}_{i} \in \set{A}({x}_{i})}
\mathbb{H}[{Z}_{\tau({x}_{i}, {a}_{i})}|Z_{{x}_{0:i}}] + V^{\pi^{\ast}}_{i+1}( ({x}_{0:i},\tau({x}_{i}, {a}_{i})) )\\
\displaystyle V^{\pi^{\ast}}_{t}({x}_{0:t}) =& \hspace{-0mm}\displaystyle\max_{{a}_{t} \in \set{A}({x}_t)}
\mathbb{H}[{Z}_{\tau({x}_{t}, {a}_{t})}| Z_{{x}_{0:t}}]
\end{array}
\label{eq:9.1}
\end{equation}
for stage $i = 0,\ldots,t-1$. 
Since each stagewise posterior entropy $\mathbb{H}[{Z}_{\tau({x}_{i}, {a}_{i})}|Z_{{x}_{0:i}}]$ (\ref{eq:7a}) can be evaluated in closed form as explained above,
$i$MASP for sampling the GP-based field (\ref{eq:9.1}) reduces to a deterministic planning problem. Furthermore, it turns out to be the well-known maximum entropy sampling problem \cite{Shewry87} as demonstrated in \cite{LowICAPS09}.
Policy $\pi^{\ast} = \langle\pi^{\ast}_0(x_{0:0}),\ldots, \pi^{\ast}_{t}(x_{0:t})\rangle$ can be determined by
\begin{equation}
\hspace{-0mm}
\begin{array}{rl}
\pi^{\ast}_{i}(x_{0:i}) =& \displaystyle\mathop{\arg\max}_{{a}_{i} \in \set{A}({x}_{i})}
\mathbb{H}[{Z}_{\tau({x}_{i}, {a}_{i})}|Z_{{x}_{0:i}}] + V^{\pi^{\ast}}_{i+1}( ({x}_{0:i},\tau({x}_{i}, {a}_{i})) )\\
\pi^{\ast}_{t}(x_{0:t}) =& \displaystyle\mathop{\arg\max}_{{a}_{t} \in \set{A}({x}_t)}
\mathbb{H}[{Z}_{\tau({x}_{t}, {a}_{t})}| Z_{{x}_{0:t}}]
\end{array}
\label{eq:9.1a}
\end{equation}
for stage $i = 0,\ldots,t-1$. 
Similar to the optimal value functions (\ref{eq:9.1}), $\pi^{\ast}$ only requires the history of robots' sampling locations as inputs. So, $\pi^{\ast}$ can generate the maximum-entropy paths prior to exploration.

Solving the myopic formulation of $i$MASP (\ref{eq:9.1}) is often considered to ease computation (Section~\ref{sect:time}), 
which entails deriving the non-Markovian greedy policy $\pi^{G} = \langle\pi^{G}_0(x_{0:0}),\ldots,$ $\pi^{G}_{t}(x_{0:t})\rangle$ where, for stage $i = 0,\ldots,t$,
\begin{equation}
\pi^{G}_{i}(x_{0:i}) = \displaystyle\mathop{\arg\max}_{{a}_{i} \in \set{A}({x}_i)}
\mathbb{H}[{Z}_{\tau({x}_{i}, {a}_{i})}| Z_{{x}_{0:i}}] \ . 
\label{eq:9.1g}
\end{equation}
The work of \cite{Guestrin08} has proposed a non-Markovian greedy policy $\pi^M = \langle\pi^{M}_0(x_{0:0}),\ldots, \pi^{M}_{t}(x_{0:t})\rangle$ to approximately maximize the closely related mutual information criterion:
\begin{equation}
\pi^{M}_{i}(x_{0:i}) = \displaystyle\mathop{\arg\max}_{{a}_{i} \in \set{A}({x}_i)}
\mathbb{H}[{Z}_{\tau({x}_{i}, {a}_{i})}| Z_{{x}_{0:i}}] - \mathbb{H}[{Z}_{\tau({x}_{i}, {a}_{i})}| Z_{\overline{x}_{0:i+1}}]
\label{eq:9.1g2}
\end{equation}
for stage $i = 0,\ldots,t$
where $\overline{{x}}_{0:i+1}$ denotes the vector comprising locations of domain $\set{U}$ not found in $({x}_{0:i},\tau({x}_{i}, {a}_{i}))$.
It is shown in \cite{Guestrin08} that $\pi^M$ greedily selects new sampling locations that maximize the increase in mutual information.
As noted in \cite{LowICAPS09}, this strategy is deficient in that it may not necessarily minimize the mapping uncertainty defined using the entropy criterion.
More importantly, it suffers a huge computational drawback: the time needed to derive $\pi^M$ depends on the map resolution (i.e., $|\set{U}|$) (Section~\ref{sect:time}).
%
%
\section{Markov-Based Path Planning}
\label{sect:mpp}
The Markov property assumes that the measurements ${Z}_{{x}_{i+1}}$ to be observed next in stage $i+1$ depends only on the current measurements $Z_{{x}_{i}}$ observed in stage $i$ and is conditionally independent of the past measurements $Z_{{x}_{0:i-1}}$ observed from stages $0$ to $i-1$.
That is, $f(z_{x_{i+1}}|z_{x_{0:i}}) = f(z_{x_{i+1}}|z_{x_{i}})$ for all $z_{x_0},z_{x_1},\ldots,z_{x_{i+1}}$.
As a result, $\mathbb{H}[{Z}_{{x}_{i+1}}| Z_{{x}_{0:i}}]$ (\ref{eq:7a}) can be approximated by $\mathbb{H}[{Z}_{{x}_{i+1}}| Z_{{x}_{i}}]$.
It is therefore straightforward to impose the Markov assumption on $i$MASP (\ref{eq:9.1}), which yields the following dynamic programming equations for the Markov-based path planning problem:
\begin{equation}
\begin{array}{rl}
\displaystyle \widetilde{V}_{i}({x}_{i}) 
=& \hspace{-0mm}\displaystyle\max_{{a}_{i} \in \set{A}({x}_{i})}
\mathbb{H}[{Z}_{\tau({x}_{i}, {a}_{i})}| Z_{{x}_{i}}] + \widetilde{V}_{i+1}(\tau({x}_{i}, {a}_{i}))\\
\displaystyle \widetilde{V}_{t}({x}_{t}) =& \hspace{-0mm}\displaystyle\max_{{a}_{t} \in \set{A}({x}_t)}
\mathbb{H}[{Z}_{\tau({x}_{t}, {a}_{t})}| Z_{{x}_{t}}] \ .
\end{array}
\label{eq:9.2}
\end{equation}
for stage $i = 0,\ldots,t-1$.
Consequently, the Markov-based policy $\widetilde{\pi} = \langle\widetilde{\pi}_0({x}_{0}),\ldots, \widetilde{\pi}_{t}({x}_{t})\rangle$
can be determined by
\begin{equation}
\begin{array}{rl}
\widetilde{\pi}_{i}({x}_{i}) =& \displaystyle\mathop{\arg\max}_{{a}_{i} \in \set{A}({x}_{i})} \mathbb{H}[{Z}_{\tau({x}_{i}, {a}_{i})}| Z_{{x}_{i}}] + \widetilde{V}_{i+1}(\tau({x}_{i}, {a}_{i}))\\ 
\widetilde{\pi}_{i}({x}_{t}) =& \displaystyle\mathop{\arg\max}_{{a}_{t} \in \set{A}({x}_t)}
\mathbb{H}[{Z}_{\tau({x}_{t}, {a}_{t})}| Z_{{x}_{t}}] \ .
\end{array}
\label{eq:9.2op}
\end{equation}
%
\subsection{\hspace{-1.3mm}Time Complexity: Analysis \& Comparison}
\label{sect:time}
\vspace{-1mm}
\begin{theorem}
Let $\set{A} \defeq \set{A}({x}_0) = \ldots = \set{A}({x}_{t})$.
Deriving the Markov-based policy $\widetilde{\pi}$ (\ref{eq:9.2op}) for the transect sampling task requires $\set{O}(|\set{A}|^2 (t+k^4))$ time.
\vspace{-1mm}
\label{thm:1}
\end{theorem}
Note that 
$|\set{A}| =$ $^r\mathrm{C}_k = \set{O}(r^k)$ where $r$ is the number of sampling locations per column and $k \leq r$ as assumed in Section~\ref{sect:tstask}.
Though $|\set{A}|$ is exponential in the number $k$ of robots, $r$ is expected to be small in a transect, which prevents $|\set{A}|$ from growing too large.

In contrast, deriving $i$MASP-based policy $\pi^{\ast}$ (\ref{eq:9.1a}) requires $\set{O}(|\set{A}|^t t^2 k^4)$ time.
Deriving greedy policies $\pi^{G}$ (\ref{eq:9.1g}) and $\pi^{M}$ (\ref{eq:9.1g2}) incur, respectively, $\set{O}(|\set{A}| t^4 k^3 + |\set{A}|^2 t k^4)$ and $\set{O}(|\set{A}| t |\set{U}|^3 + |\set{A}|^2 t k^4)=\set{O}(|\set{A}| t^4 r^3 + |\set{A}|^2 t k^4)$ time to compute the observation paths over all $|\set{A}|$ possible choices of starting robot locations. Clearly, all the non-Markovian strategies do not scale as well as our Markov-based approach with increasing length $t+1$ of planning horizon or number $t+2$ of columns, which is expected to be large.
As demonstrated empirically (Section~\ref{sect:expt}), the Markov-based policy $\widetilde{\pi}$ can be derived faster than $\pi^G$ and $\pi^{M}$ by more than an order of magnitude; this computational advantage is boosted further for transect sampling tasks with unknown starting robot locations.
\subsection{Performance Guarantees}
\label{sect:guarantee}
We will first provide a theoretical guarantee on how the Markov-based policy $\widetilde{\pi}$ (\ref{eq:9.2op}) performs relative to the non-Markovian $i$MASP-based policy $\pi^\ast$ (\ref{eq:9.1a}) for the case of $1$ robot. 
This key result follows from our intuition that when the horizontal spatial correlation becomes small, exploiting the past measurements for path planning should hardly improve the active sampling performance in a transect sampling task,
thus favoring the Markov-based policy.
Though this intuition is simple, supporting it with formal theoretical results and their corresponding proofs (Appendix~\ref{sect:proofs}) turns out to be non-trivial as shown below. 

Recall the Markov assumption that $\mathbb{H}[{Z}_{{x}_{i+1}}|Z_{{x}_{0:i}}]$ (\ref{eq:7a}) is to be approximated by $\mathbb{H}[{Z}_{{x}_{i+1}}|Z_{{x}_{i}}]$. This prompts us to first consider bounding the difference of these posterior entropies that ensues from the Markov property:
\vspace{-2mm}
\begin{equation}
\begin{array}{l}
\mathbb{H}[{Z}_{{x}_{i+1}}| Z_{{x}_{i}}] - \mathbb{H}[{Z}_{{x}_{i+1}}| Z_{{x}_{0:i}}] = \displaystyle \frac{1}{2} \log\frac{\sigma^2_{{{x}_{i+1}}\mid {x}_{i}}}{\sigma^2_{{{x}_{i+1}}\mid {x}_{0:i}}}\vspace{1mm}\\
\hspace{10mm}= \displaystyle \frac{1}{2} \log \left( 1 - \frac{\sigma^2_{{{x}_{i+1}}\mid {x}_{i}}-\sigma^2_{{{x}_{i+1}}\mid {x}_{0:i}}}{\sigma^2_{{{x}_{i+1}}\mid {x}_{i}}} \right)^{-1} \geq 0 \ .
\end{array}
\label{eq:9.6}
\end{equation}
This difference can be interpreted as the reduction in uncertainty of the measurements ${Z}_{{x}_{i+1}}$ to be observed next in stage $i+1$ by observing the past measurements $Z_{{x}_{0:i-1}}$ from stages $0$ to $i-1$ given the current measurements $Z_{{x}_{i}}$ observed in stage $i$.
This difference is small if $Z_{{x}_{0:i-1}}$ does not contribute much to the reduction in uncertainty of ${Z}_{{x}_{i+1}}$ given $Z_{{x}_{i}}$.
It (\ref{eq:9.6}) is often known as the conditional mutual information of ${Z}_{{x}_{i+1}}$ and $Z_{{x}_{0:i-1}}$ given $Z_{{x}_{i}}$ denoted by
$$\mathbb{I}[{Z}_{{x}_{i+1}};Z_{{x}_{0:i-1}}|Z_{{x}_{i}}] \defeq \mathbb{H}[{Z}_{{x}_{i+1}}| Z_{{x}_{i}}] - \mathbb{H}[{Z}_{{x}_{i+1}}| Z_{{x}_{0:i}}] \ , $$
which is of value $0$ if the Markov property holds. 

The results to follow assume that the transect is discretized into a grid of sampling locations.
Let $\omega_1$ and $\omega_2$ denote the horizontal and vertical grid discretization widths (i.e., separations between adjacent sampling locations), respectively.
Let ${\ell}'_1 \defeq \ell_1/\omega_1$ and ${\ell}'_2 \defeq \ell_2/\omega_2$ represent the normalized horizontal and vertical length-scale components, respectively.
The following lemma bounds the variance reduction term $\sigma^2_{{{x}_{i+1}}\mid {x}_{i}}-\sigma^2_{{{x}_{i+1}}\mid {x}_{0:i}}$ in (\ref{eq:9.6}):
\vspace{-1mm}
\begin{lemma} 
Let $\displaystyle\xi\hspace{-0mm}\defeq\hspace{-0mm}\exp\left\{-\frac{1}{2{\ell}'^2_1} \right\}$ and $\rho\hspace{-0mm}\defeq\hspace{-0mm}\displaystyle 1 + \frac{\sigma^2_n}{\sigma^2_s}$.
If $\displaystyle\xi\hspace{-0mm}<\hspace{-0mm}\frac{\rho}{i}$, then $\displaystyle0\hspace{-0mm}\leq\hspace{-0mm}\sigma^2_{{{x}_{i+1}}\mid {x}_{i}} - \ \sigma^2_{{{x}_{i+1}}\mid {x}_{0:i}}\hspace{-0mm}\leq\hspace{-0mm}\frac{\sigma^2_s \xi^4}{\frac{\rho}{i}  - \xi} .$
\label{lem:9.7}
\end{lemma}
The next lemma is fundamental to the subsequent results on the active sampling performance of Markov-based policy $\widetilde{\pi}$. It provides bounds on 
$\mathbb{I}[{Z}_{{x}_{i+1}};Z_{{x}_{0:i-1}}|Z_{{x}_{i}}]$,
which follow immediately from (\ref{eq:9.6}), Lemma~\ref{lem:9.7}, and the lower bound 
$$
\sigma^2_{{{x}_{i+1}}\mid {x}_{i}} = 
\displaystyle\sigma^2_{{{x}_{i+1}}} - (\sigma_{{{x}_{i+1}} {{x}_{i}}})^2/\sigma^2_{{{x}_{i}}} \geq \displaystyle\sigma^2_s + \sigma^2_n - \sigma^2_s\xi^2 :
$$
\begin{lemma} 
If $\displaystyle\xi\hspace{-0.15mm} <\hspace{-0.15mm} \frac{\rho}{i}$,
then $0\hspace{-0.15mm}\leq\hspace{-0.15mm}\mathbb{I}[{Z}_{{x}_{i+1}};Z_{{x}_{0:i-1}}|Z_{{x}_{i}}]
\hspace{-0.15mm}\leq\hspace{-0.15mm}\Delta(i)
$
\hspace{-0mm}where\hspace{-0mm}
$\displaystyle\Delta(i)\hspace{-0mm}\defeq\hspace{-0mm}
\displaystyle \frac{1}{2} \log\hspace{-0mm}\left(\hspace{-0mm}1 - \frac{\xi^4}{(\frac{\rho}{i}  - \xi)(\rho -\xi^2)}\hspace{-0mm} \right)^{\hspace{-0mm}-1}\hspace{-0mm}.$
\vspace{1mm}

\noindent
{\scshape{Remark}}. If $j\leq s$, then $\Delta(j)\leq\Delta(s)$ for $j,s = 0,\ldots,t$.
\label{lem:9.8}
\end{lemma}
From Lemma~\ref{lem:9.8}, since $\Delta(i)$ bounds $\mathbb{I}[{Z}_{{x}_{i+1}};Z_{{x}_{0:i-1}}|Z_{{x}_{i}}]$ from above, 
a small $\mathbb{I}[{Z}_{{x}_{i+1}};Z_{{x}_{0:i-1}}|Z_{{x}_{i}}]$ can be guaranteed by making $\Delta(i)$ small.
From the definition of $\Delta(i)$, there are a few ways to achieve a small $\Delta(i)$: 
(a) $\Delta(i)$ depends on ${\ell}'_1$ through $\xi$.
As ${\ell}'_1\rightarrow 0^{+}$, $\xi\rightarrow 0^{+}$, by definition. Consequently, $\Delta(i)\rightarrow 0^{+}$.
A small ${\ell}'_1$ can be obtained using a small ${\ell}_1$ and/or a large $\omega_1$, by definition;
(b) $\Delta(i)$ also depends on the noise-to-signal ratio $\sigma^2_n/\sigma^2_s$ through $\rho$.
Raising $\sigma^2_n$ or lowering $\sigma^2_s$ increases $\rho$, by definition. This, in turn, decreases $\Delta(i)$;
(c) Since $i$ indicates the length of history of observations,
the remark after Lemma~\ref{lem:9.8} tells us that a shorter length produces a smaller $\Delta(i)$.
To summarize, (a) environmental field conditions such as 
smaller horizontal spatial correlation and noisy, less intense fields, and (b) 
sampling task settings such as larger horizontal grid discretization width and shorter length of history of observations
all contribute to smaller $\Delta(i)$, and hence smaller $\mathbb{I}[{Z}_{{x}_{i+1}};Z_{{x}_{0:i-1}}|Z_{{x}_{i}}]$. 
This analysis is important for understanding the practical implication of our theoretical results later.
A limitation with using Lemma~\ref{lem:9.8} is that of the sufficient condition $\xi < \rho/i$, which will hold if the field conditions and task settings realized above to make $\Delta(i)$ small are adequately satisfied.

The following theorem uses the induced optimal value $\widetilde{V}_{0}(x_{0})$ from solving the Markov-based path planning problem (\ref{eq:9.2}) to bound the maximum entropy ${V}^{\pi^\ast}_{0}({x}_{0})$ of observation paths achieved by $\pi^\ast$ from solving $i$MASP (\ref{eq:9.1}):
\begin{theorem}
Let $\epsilon_i \defeq \sum^t_{s=i} \Delta(s) \leq (t-i+1) \Delta(t)$. If $\displaystyle\xi\hspace{-0mm} <\hspace{-0mm} \frac{\rho}{t}$, then
$\widetilde{V}_{i}({x}_{i}) - \epsilon_i \leq {V}^{\pi^\ast}_{i}({x}_{0:i}) \leq \widetilde{V}_{i}({x}_{i})$ for $i = 0,\ldots,t$.
\vspace{-3mm}
\label{thm:9.9}
\end{theorem}
The above result is useful in providing an efficient way of knowing the maximum entropy ${V}^{\pi^\ast}_{0}({x}_{0})$, albeit approximately:
the time needed to derive the two-sided bounds on ${V}^{\pi^\ast}_{0}({x}_{0})$ is linear in the length of planning horizon (Theorem~\ref{thm:1}) as opposed to exponential time required to compute the exact value of ${V}^{\pi^\ast}_{0}({x}_{0})$.
Since the error bound $\epsilon_i$ is defined as a sum of $\Delta(s)$'s, we can rely on the above analysis of $\Delta(s)$ (see paragraph after Lemma~\ref{lem:9.8}) to  improve this error bound: (a) environmental field conditions such as
smaller horizontal spatial correlation and noisy, less intense fields, and (b) sampling task settings such as larger horizontal grid discretization width and shorter planning horizon (i.e., fewer transect columns) all improve this error bound.

In the main result below, the Markov-based policy $\widetilde{\pi}$ is guaranteed to achieve an entropy $V^{\widetilde{\pi}}_{0}({x}_{0})$ of observation paths (i.e., by plugging $\widetilde{\pi}$ into (\ref{eq:9.9a})) that is not more than $\epsilon_0$ 
from the maximum entropy $V^{\pi^\ast}_{0}({x}_0)$ of observation paths achieved by policy $\pi^\ast$:
\vspace{-1mm}
\begin{theorem}
If $\displaystyle\xi < \frac{\rho}{t}$, then policy $\widetilde{\pi}$ is $\epsilon_0$-optimal in achieving the maximum-entropy criterion,
i.e., $V^{\pi^\ast}_{0}({x}_0) - V^{\widetilde{\pi}}_{0}({x}_{0})\leq\epsilon_0$.\vspace{-1mm}
\label{thm:9.10}
\end{theorem}
Again, since the error bound $\epsilon_0$ is defined as a sum of $\Delta(s)$'s, we can use the above analysis of $\Delta(s)$ to improve this bound: 
(a) environmental field conditions such as
smaller horizontal spatial correlation and noisy, less intense fields, and (b) sampling task settings such as larger horizontal grid discretization width and shorter planning horizon (i.e., fewer transect columns) all result in smaller $\epsilon_0$, and hence improve the active sampling performance of Markov-based policy $\widetilde{\pi}$ relative to that of non-Markovian $i$MASP-based policy $\pi^\ast$.
This not only supports our prior intuition (see first paragraph of this section) but
also identifies other means of limiting the performance degradation of the Markov-based policy.

For the multi-robot case, a condition has to be imposed on the covariance structure of GP to obtain a similar guarantee:
\begin{equation}
|\sigma_{uv|x_{0:i}}| \leq |\sigma_{uv|x_m}|\vspace{-0mm} 
\label{eq:9.10}
\end{equation}
for $m=0,\ldots,i$ and any $u,v,x_0,x_1,\ldots,x_i\in\set{U}$. 
Intuitively, (\ref{eq:9.10}) says that further conditioning does not make $Z_u$ and $Z_v$ more correlated. Note that (\ref{eq:9.10}) is satisfied if $u=v$. 

Similar to Lemma~\ref{lem:9.8} for the $1$-robot case, we can bound 
$\mathbb{I}[{Z}_{{x}_{i+1}};Z_{{x}_{0:i-1}}|Z_{{x}_{i}}]$
for the multi-robot case but tighter conditions have to be satisfied:
\begin{lemma} 
Let ${\ell}'_1={\ell}'_2$. If 
$\displaystyle \xi < \min(\frac{\rho}{ik},\frac{\rho}{4k})$ 
and \emph{(\ref{eq:9.10})} is satisfied,
then $\displaystyle
0 \leq 
\mathbb{I}[{Z}_{{x}_{i+1}};Z_{{x}_{0:i-1}}|Z_{{x}_{i}}]
\leq \Delta_k(i)$
where
$\displaystyle\Delta_k(i) \defeq 
\displaystyle \frac{k}{2} \log \left( 1 - \frac{\xi^4}{(\frac{\rho}{ik}  - \xi)(\rho -\frac{4k}{\rho}\xi^2)} \right)^{-1}.$
\label{lem:9.11}
\end{lemma}
To improve the upper bound $\Delta_k(i)$, the above analysis of $\Delta(i)$ 
can be applied here as these two upper bounds are largely similar:
(a) environmental field conditions such as smaller spatial correlation and noisy, less intense fields, and (b) sampling task settings such as larger grid discretization width  and shorter planning horizon (i.e., fewer transect columns) all entail smaller $\Delta_k(i)$. Decreasing the number $k$ of robots also reduces $\Delta_k(i)$, thus yielding tighter bounds on $\mathbb{I}[{Z}_{{x}_{i+1}};Z_{{x}_{0:i-1}}|Z_{{x}_{i}}]$. Using Lemma~\ref{lem:9.11}, we can derive guarantees similar to that of Theorems~\ref{thm:9.9} and~\ref{thm:9.10} on the performance of Markov-based policy $\widetilde{\pi}$ for the multi-robot case.
\section{Experiments and Discussion}
\label{sect:expt}
In Section~\ref{sect:guarantee}, we have highlighted the practical implication of our main theoretical result (i.e., Theorem~\ref{thm:9.10}), which establishes various environmental field conditions and sampling task settings to limit the performance degradation of Markov-based policy $\widetilde{\pi}$. 
This result, however, does not reveal whether $\widetilde{\pi}$ performs well (or not) under ``seemingly'' less favorable field conditions and task settings that do not jointly satisfy its sufficient condition $\xi<\rho/(tk)$.
These include
large spatial correlation, less noisy, highly intense fields, small grid discretization width, long planning horizon (i.e., many transect columns), and large number of robots.
So, this section evaluates the active sampling performance and time efficiency of $\widetilde{\pi}$ empirically on two real-world datasets under such field conditions and task settings as detailed below: (a) May~$2009$ temperature field data of Panther Hollow Lake in Pittsburgh, PA spanning $25$~m by $150$~m, and (b) June~$2009$ plankton density field data of Chesapeake Bay spanning $314$~m by $1765$~m. 

Using maximum likelihood estimation (MLE) \cite{Rasmussen06}, the learned hyperparameters (i.e., horizontal and vertical length-scales, signal and noise variances) are, respectively, $\ell_1 =40.45$~m, $\ell_2 = 16.00$~m, $\sigma^2_s =0.1542$, and $\sigma^2_n = 0.0036$
for the temperature field, and $\ell_1 = 27.53$~m, $\ell_2 = 134.64$~m, $\sigma^2_s =2.152$, and $\sigma^2_n = 0.041$ for the plankton density field. It can be observed that the temperature and plankton density fields have low noise-to-signal ratios $\sigma^2_n/\sigma^2_s$ of $0.023$ and $0.019$, respectively. Relative to the size of transect, 
both fields have large vertical spatial correlations, but only the temperature field has large horizontal spatial correlation.

The performance of Markov-based policy $\widetilde{\pi}$ is compared to non-Markovian policies produced by two state-of-the-art information-theoretic exploration strategies: greedy policies $\pi^G$ (\ref{eq:9.1g}) and $\pi^M$ (\ref{eq:9.1g2}) proposed by \cite{LowICAPS09} and \cite{Guestrin08}, respectively. 
The non-Markovian policy $\pi^\ast$ that has to be derived approximately using Learning Real-Time A$^{\ast}$ is excluded from comparison due to the reason provided in Section~\ref{sect:intro}.
%

\subsection{Performance Metrics} 
The tested policies are evaluated using the two metrics proposed in \cite{LowICAPS09}, which quantify the mapping uncertainty of the unobserved areas of the field differently:
(a) The ENT($\pi$) metric measures the posterior joint entropy $\mathbb{H}[Z_{\overline{{x}}_{0:t+1}}|Z_{x_{0:t+1}}]$ of field measurements $Z_{\overline{{x}}_{0:t+1}}$ at unobserved locations $\overline{{x}}_{0:t+1}$ 
where $\overline{{x}}_{0:t+1}$ denotes the vector comprising locations of domain $\set{U}$ not found in the sampled locations ${{x}}_{0:t+1}$ selected by policy $\pi$.
Smaller ENT($\pi$) implies lower mapping uncertainty;
(b) The ERR($\pi$) metric measures the mean-squared relative error $|\set{U}|^{-1} \sum_{u\in \set{U}} \{(z_{u} - \mu_{{u}|x_{0:t+1}})/\bar{\mu}\}^2$ resulting from using the observations (i.e., sampled locations ${{x}}_{0:t+1}$ and corresponding measurements $z_{{{x}}_{0:t+1}}$) selected by policy $\pi$ 
and the posterior mean $\mu_{{u}|x_{0:t+1}}$ (\ref{eq:6}) 
to predict the field where $\bar{\mu} = |\set{U}|^{-1} \sum_{u\in \set{U}} z_{u}$.
Smaller ERR($\pi$) implies higher prediction accuracy.
Two noteworthy differences distinguish these metrics: (a) The ENT($\pi$) metric exploits the spatial correlation between field measurements in the unobserved areas
whereas the ERR($\pi$) metric implicitly assumes independence between them.
As a result, 
unlike the ERR($\pi$) metric, the ENT($\pi$) metric does not overestimate the mapping uncertainty.
To illustrate this, suppose the unknown field measurements are restricted to only two unobserved locations $u$ and $v$ residing in a highly uncertain area and they are highly correlated due to spatial proximity.
The behavior of the ENT($\pi$) metric can be understood upon applying the chain rule for entropy 
(i.e., ENT($\pi$) $= \mathbb{H}[Z_u, Z_v|Z_{x_{0:t+1}}] = \mathbb{H}[Z_u|Z_{x_{0:t+1}}] + \mathbb{H}[Z_v|Z_{x_{0:t+1}}, Z_u]$); the latter uncertainty term (i.e., posterior entropy of $Z_v$) is significantly reduced or ``discounted'' due to the high spatial correlation between $Z_u$ and $Z_v$. Hence, the mapping uncertainty of these two unobserved locations is not overestimated.
A practical advantage of this metric is that it does not overcommit sensing resources; in the simple illustration above, a single observation at either location $u$ or $v$ suffices to learn both field measurements well.
On the other hand, the ERR($\pi$) metric considers each location to be of high uncertainty due to the independence assumption;
(b) In contrast to the ENT($\pi$) metric, the ERR($\pi$) metric can use ground truth measurements to evaluate if the field is being mapped accurately.
Let ENTD($\pi$) $\defeq$ ENT($\widetilde{\pi}$)$-$ENT($\pi$) and 
ERRD($\pi$) $\defeq$ ERR($\widetilde{\pi}$)$-$ERR($\pi$).
Decreasing ENTD($\pi$) improves the ENT($\widetilde{\pi}$) performance of $\widetilde{\pi}$ relative to that of $\pi$.
Small $|$ENTD($\pi$)$|$ implies that $\widetilde{\pi}$ achieves ENT($\widetilde{\pi}$) performance comparable to that of $\pi$.
ERRD($\pi$) can be interpreted likewise.
Additionally, we will consider the time taken to derive each policy as the third metric.
\subsection{Temperature Field Data}
\label{sect:tfield}
\begin{figure}
\begin{tabular}{c}
\hspace{-3mm}\epsfig{figure=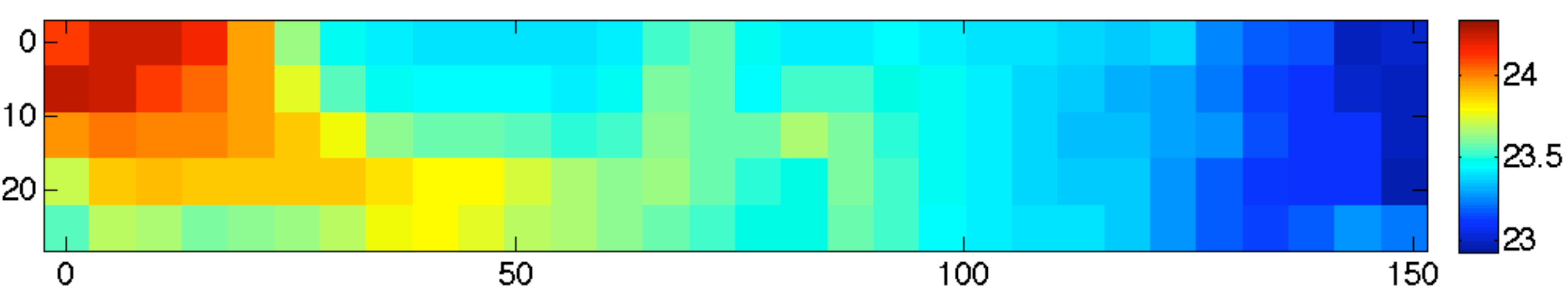,width=8.7cm}\vspace{-1mm}\\
\hspace{-3mm}{(a) Field $1$: $\ell_1 = 5.00$~m, $\ell_2 = \ \ 5.00$~m.}\\
\hspace{-3mm}\epsfig{figure=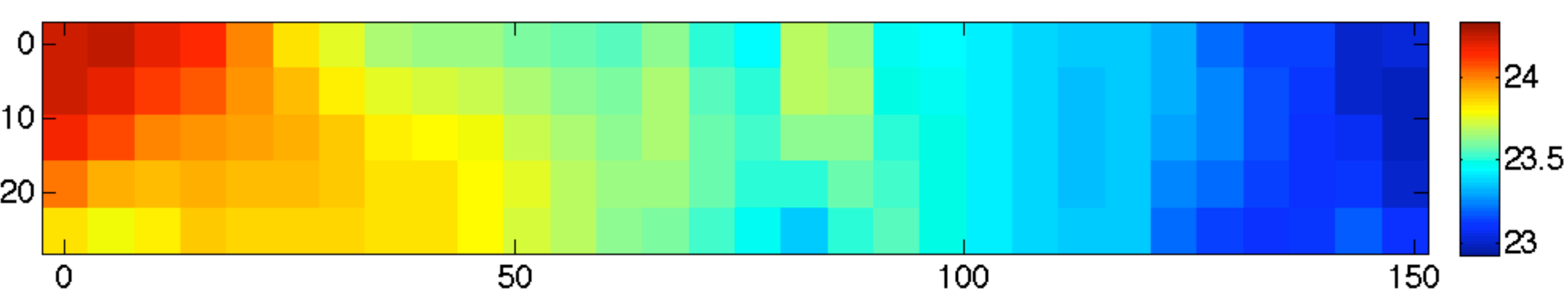,width=8.7cm}\vspace{-1mm}\\
\hspace{-3mm}{(b) Field $2$: $\ell_1 = 5.00$~m, $\ell_2 = 16.00$~m.}\\
\hspace{-3mm}\epsfig{figure=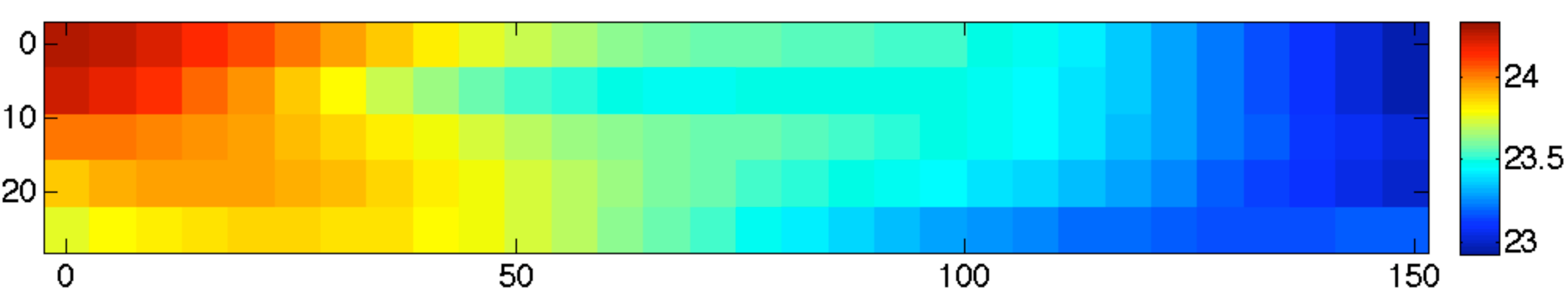,width=8.7cm}\vspace{-1mm}\\
\hspace{-3mm}{(c) Field $3$: $\ell_1 = 40.45$~m, $\ell_2 = \ \ 5.00$~m.}\vspace{0mm}\\
\hspace{-3mm}\epsfig{figure=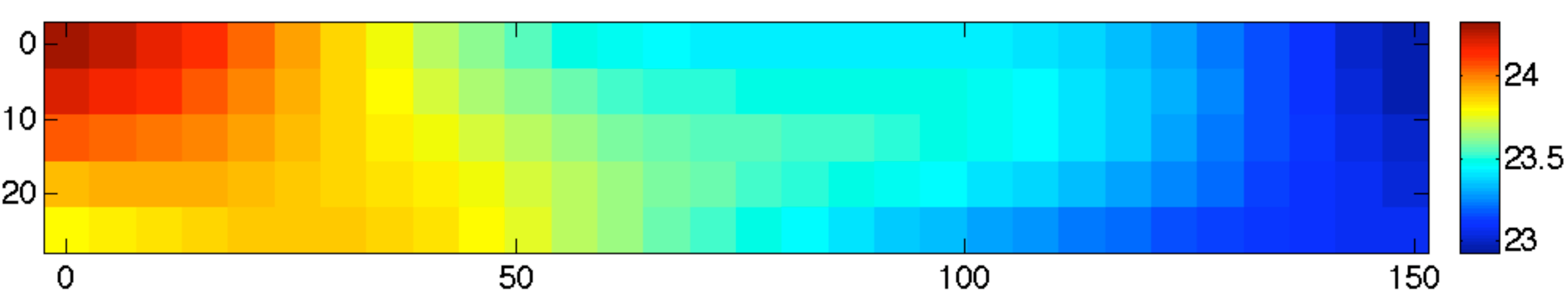,width=8.7cm}\vspace{-1mm}\\
\hspace{-3mm}{(d) Field $4$: $\ell_1 = 40.45$~m, $\ell_2 = 16.00$~m.}\vspace{-2mm}
\end{tabular}
\caption{Temperature fields (measured in$\,^{\circ}\mathrm{C}$) with varying horizontal length-scale $\ell_1$ and vertical length-scale $\ell_2$.}
\vspace{-5mm}
\label{fig:9.2}
\end{figure}
We will first investigate how varying spatial correlations (i.e., varying length-scales) of the temperature field affect the ENT($\pi$) and ERR($\pi$) performance of evaluated policies.
The temperature field is discretized into a $5 \times 30$ grid of sampling locations as shown in Figs.~\ref{fig:tempgrid} and~\ref{fig:9.2}d.
The horizontal and/or vertical length-scales of the original field (i.e., field $4$ in Fig.~\ref{fig:9.2}d) are reduced to produce modified fields $1$, $2$, and $3$ (respectively, Figs.~\ref{fig:9.2}a,~\ref{fig:9.2}b, and~\ref{fig:9.2}c); we fix these reduced length-scales while learning the remaining hyperparameters (i.e., signal and noise variances) through MLE.

Table~\ref{tab:entcompare}
shows the results of mean ENT($\pi$) and ERR($\pi$) performance of tested policies (i.e., averaged over all possible starting robot locations) with varying length-scales and number of robots. 
The ENT($\pi$) and ERR($\pi$) for all policies generally decrease with increasing length-scales (except ERR($\widetilde{\pi}$) for $1$ robot from field $2$ to $4$) due to increasing spatial correlation between measurements, thus resulting in lower mapping uncertainty. 
\begin{table}
\caption{Comparison of ENT($\pi$) (left) and ERR($\pi$)  ($\times 10^{-5}$) (right) performance for temperature fields that are discretized into $5 \times 30$ grids (Fig.~\ref{fig:9.2}).}
\label{tab:entcompare}
\begin{tiny}
\hspace{-2mm}
\begin{tabular}{cc}
\hspace{-8.5mm}
\begin{tabular}{|c|c|c|c|c|}
\hline
1 robot\hspace{1mm} & \multicolumn{4}{|c|}{Field}\\
\hline
Policy & 1 & 2 & 3 & 4 \\
\hline
$\widetilde{\pi}$ & -83 & -246 & -543 & -597 \\
$\pi^G$ & -82 & -246 & -554 & -598 \\
$\pi^{M}$ & -80 & -211 & -554 & -596 \\
\hline
\end{tabular}
&\hspace{-5.3mm}
\begin{tabular}{|c|c|c|c|c|}
\hline
1 robot\hspace{1mm} & \multicolumn{4}{|c|}{Field}\\
\hline
Policy & 1 & 2 & 3 & 4 \\
\hline
$\widetilde{\pi}$ & 3.7040 & 0.5713 & 2.3680 & 0.5754 \\
$\pi^G$ & 1.8680 & 0.5713 & 0.0801 & 0.0252 \\
$\pi^{M}$ & 1.8433 & 0.5212 & 0.0701 & 0.0421 \\
\hline
\end{tabular}\\

\hspace{-8.5mm}
\begin{tabular}{|c|c|c|c|c|}
\hline
2 robots & \multicolumn{4}{|c|}{Field}\\
\hline
Policy & 1 & 2 & 3 & 4 \\
\hline
$\widetilde{\pi}$ & -71 & -190 & -380 & -422 \\
$\pi^G$ & -72 & -190 & -382 & -425 \\
$\pi^{M}$ & -68 & -131 & -382 & -421 \\
\hline
\end{tabular}
&
\hspace{-5.3mm}
\begin{tabular}{|c|c|c|c|c|}
\hline
2 robots & \multicolumn{4}{|c|}{Field}\\
\hline
Policy & 1 & 2 & 3 & 4 \\
\hline
$\widetilde{\pi}$ & 0.3797 & 0.2101 & 0.1171 & 0.0095 \\
$\pi^G$ & 0.3526 & 0.2101 & 0.0150 & 0.0087 \\
$\pi^{M}$ & 0.6714 & 0.1632 & 0.0148 & 0.0086 \\
\hline
\end{tabular}\\
\hspace{-8.5mm}
\begin{tabular}{|c|c|c|c|c|}
\hline
3 robots & \multicolumn{4}{|c|}{Field}\\
\hline
Policy & 1 & 2 & 3 & 4 \\
\hline
$\widetilde{\pi}$ & -53 & -109 & -232 & -297 \\
$\pi^G$ & -53 & -109 & -215 & -297 \\
$\pi^{M}$ & -53 & -73 & -214 & -255 \\
\hline
\end{tabular}
& \hspace{-4.5mm}
\begin{tabular}{|c|c|c|c|c|}
\hline
3 robots & \multicolumn{4}{|c|}{Field}\\
\hline
Policy & 1 & 2 & 3 & 4 \\
\hline
$\widetilde{\pi}$ & 0.1328 & 0.0068 & 0.0063 & 0.0031 \\
$\pi^G$ & 0.1312 & 0.0068 & 0.0059 & 0.0031 \\
$\pi^{M}$ & 0.1080 & 0.1397 & 0.0055 & 0.0030 \\
\hline
\end{tabular}
\vspace{-4mm}
\end{tabular}
\end{tiny}
\end{table}

For the case of $1$ robot, the observations are as follows:
(a) When $\ell_2$ is kept constant (i.e., at $5$~m or $16$~m), reducing $\ell_1$ from $40.45$~m to $5$~m (i.e., from field $3$ to $1$ or field $4$ to $2$) decreases 
ENTD($\pi^G$), ERRD($\pi^G$),
ENTD($\pi^M$), and ERRD($\pi^M$):
when the horizontal correlation becomes small, it can no longer be exploited by the non-Markovian policies $\pi^G$ and $\pi^M$;
(b) For field $3$ with large $\ell_1$ and small $\ell_2$, ENTD($\pi^G$) and ENTD($\pi^M$) are large as the Markov property of $\widetilde{\pi}$ prevents it from exploiting the large horizontal correlation;
%
(c) When $\ell_1$ is kept constant (i.e., at $5$~m or $40.45$~m), reducing $\ell_2$ from $16$~m to $5$~m (i.e., from field $2$ to $1$ or field $4$ to $3$)
increases ERRD($\pi^G$) and ERRD($\pi^M$): when vertical correlation becomes small, it can no longer be exploited by $\widetilde{\pi}$, thus incurring larger ERR($\widetilde{\pi}$).

For the case of $2$ robots, the observations are as follows:
(a) $|$ENTD($\pi^G$)$|$ and $|$ENTD($\pi^M$)$|$ are small for all fields except for field $2$ where $\widetilde{\pi}$ significantly outperforms $\pi^M$. In particular, when $\ell_2$ is kept constant (i.e., at $5$~m or $16$~m), reducing $\ell_1$ from $40.45$~m to $5$~m (i.e., from field $3$ to $1$ or field $4$ to $2$) decreases ENTD($\pi^G$), ENTD($\pi^M$), and ERRD($\pi^G$): this is explained in the first observation of $1$-robot case; 
%
(b) For field $3$ with large $\ell_1$ and small $\ell_2$, ERRD($\pi^G$) and ERRD($\pi^M$) are large: this is explained in the second and third observations of $1$-robot case;
(c) When $\ell_1$ is kept constant (i.e., at $5$~m or $40.45$~m), reducing $\ell_2$ from $16$~m to $5$~m (i.e., from field $2$ to $1$ or field $4$ to $3$) increases ERRD($\pi^G$): this is explained in the third observation of $1$-robot case. This also holds for ERRD($\pi^M$) when $\ell_1$ is large.

For the case of $3$ robots, 
it can be observed that $\widetilde{\pi}$ can achieve ENT($\widetilde{\pi}$) and ERR($\widetilde{\pi}$) performance comparable to (if not, better than) that of $\pi^G$ and $\pi^M$ for all fields.
\begin{table}
\caption{Comparison of ENT($\pi$) (left) and ERR($\pi$)  ($\times 10^{-5}$) (right) performance for temperature field that is discretized into $13 \times 75$ grid.}
\label{tab:th}
\begin{tiny}
\begin{tabular}{cc}
\hspace{-3mm}
\begin{tabular}{|c|c|c|c|}
\hline
ENT($\pi$) & \multicolumn{3}{|c|}{Number $k$ of robots}\\
\hline
Policy & 1 & 2 & 3 \\
\hline
$\widetilde{\pi}$ & -4813 & -4284 & -3828 \\
$\pi^G$ & -4813 & -4286 & -3841 \\
$\pi^{M}$ & -4808 & -4277 & -3825 \\
\hline
\end{tabular}
&\hspace{-4mm}
\begin{tabular}{|c|c|c|c|}
\hline
ERR($\pi$) & \multicolumn{3}{|c|}{Number $k$ of robots}\\
\hline
Policy & 1 & 2 & 3 \\
\hline
$\widetilde{\pi}$ & 1.0287 & 0.0032 & 0.0015 \\
$\pi^G$ & 0.0082 & 0.0030 & 0.0024 \\
$\pi^{M}$ & 0.0087 & 0.0034 & 0.0019 \\
\hline
\end{tabular}
\end{tabular}
\vspace{-3mm}
\end{tiny}
\end{table}

To summarize the above observations on spatial correlation conditions favoring $\widetilde{\pi}$ over $\pi^G$ and $\pi^M$, $\widetilde{\pi}$ can achieve ENT($\widetilde{\pi}$) performance comparable to (if not, better than) that of $\pi^G$ and $\pi^M$ for all fields with any number of robots except for field $3$ (i.e., of large $\ell_1$ and small $\ell_2$) with $1$ robot as explained previously.
Policy $\widetilde{\pi}$ can achieve comparable ERR($\widetilde{\pi}$) performance for field $2$ (i.e., of small $\ell_1$ and large $\ell_2$) with $1$ robot because $\widetilde{\pi}$ is capable of exploiting the large vertical correlation, and the
small horizontal correlation cannot be exploited by $\pi^G$ and $\pi^M$.
Policy $\widetilde{\pi}$ can also achieve comparable ERR($\widetilde{\pi}$) performance for all fields with $2$ and $3$ robots except for field $3$ (i.e., of large $\ell_1$ and small $\ell_2$) with 2 robots.
These observations reveal that (a) small horizontal and large vertical correlations are favorable to $\widetilde{\pi}$; (b) though large horizontal and small vertical correlations are not favorable to $\widetilde{\pi}$, this problem can be mitigated by increasing the number of robots. 
For more detailed analysis (e.g., visualization of planned observation paths and their corresponding error maps), the interested reader is referred to \cite{LowPhDThesis09}.

We will now examine how the increase in resolution to $13 \times 75$ grid affects the ENT(${\pi}$) and ERR(${\pi}$) performance of evaluated policies; the resulting grid discretization width and planning horizon are about $0.4\times$ smaller and $2.5\times$ longer, respectively.
Table~\ref{tab:th} shows the results of mean ENT($\pi$) and ERR($\pi$) performance of tested policies with varying number of robots, from which we can derive observations similar to that for temperature field $4$ discretized into $5 \times 30$ grid: $\widetilde{\pi}$ can achieve ENT($\widetilde{\pi}$) and ERR($\widetilde{\pi}$) performance comparable to (if not, better than) that of $\pi^G$ and $\pi^M$ except for ERR($\widetilde{\pi}$) performance with $1$ robot.
So, increasing the grid resolution does not seem to noticeably degrade the active sampling performance of $\widetilde{\pi}$ relative to that of $\pi^G$ and $\pi^M$.

\subsection{Plankton Density Field Data}
\begin{figure}
\epsfig{figure=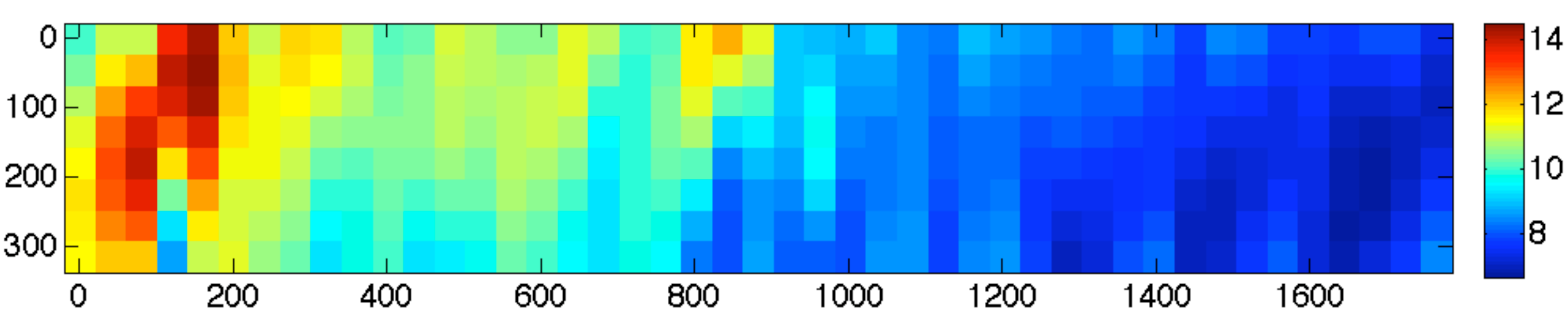,width=8.5cm}\vspace{-4mm}
\caption{Plankton density (chl-a) field (measured in $mg\ m^{-3}$) spatially distributed over a $314$~m~$\times$~$1765$~m transect that is discretized into a $8\times 45$ grid with $\ell_1 = 27.53$~m and $\ell_2 = 134.64$~m.}
\vspace{-3mm}
\label{fig:9.3}
\end{figure}
Fig.~\ref{fig:9.3} illustrates the plankton density field that is discretized into a $8\times 45$ grid.
Table~\ref{tab:cl} shows the results of mean ENT($\pi$) and ERR($\pi$) performance of tested policies with varying number of robots.
The observations are as follows:
$\widetilde{\pi}$ can achieve the same ENT($\widetilde{\pi}$) and ERR($\widetilde{\pi}$) performance as that of $\pi^G$ and superior ENT($\widetilde{\pi}$) performance over that of $\pi^M$ because small horizontal and large vertical correlations favor $\widetilde{\pi}$ as explained in Section~\ref{sect:tfield}.
By increasing the number of robots (i.e., $k> 2$), $\widetilde{\pi}$ can achieve ERR($\widetilde{\pi}$) performance comparable to (if not, better than) that of $\pi^M$.

Table~\ref{tab:ch} shows the results of mean ENT($\pi$) and ERR($\pi$) performance of tested policies after increasing the resolution to $16\times 89$ grid;
the resulting grid discretization width and planning horizon are about $0.5\times$ smaller and $2\times$ longer, respectively. Similar observations can be obtained: $\widetilde{\pi}$ can achieve ENT($\widetilde{\pi}$) performance comparable to that of $\pi^G$ and superior ENT($\widetilde{\pi}$) performance over that of $\pi^M$.
By deploying more than $1$ robot, $\widetilde{\pi}$ can achieve ERR($\widetilde{\pi}$) performance comparable to (if not, better than) that of $\pi^G$ and $\pi^M$.
Again, we can observe that increasing the grid resolution does not seem to noticeably degrade the active sampling performance of $\widetilde{\pi}$ relative to that of $\pi^G$ and $\pi^M$.
\subsection{Incurred Policy Time}
Fig.~\ref{fig:9.4} shows the time taken to derive the tested policies for sampling the temperature and plankton density fields with varying number of robots and grid resolutions. 
It can be observed that the time taken to derive $\widetilde{\pi}$ is shorter than
that needed to derive $\pi^G$ and $\pi^M$ by more than $1$ and $4$ orders of magnitude, respectively.
It is important to point out that Fig.~\ref{fig:9.4} reports the average time taken to derive $\pi^G$ and $\pi^M$ over all possible starting robot locations. So, if the starting robot locations are unknown, the incurred time to derive $\pi^G$ and $\pi^M$ have to be increased by 
$^r\mathrm{C}_k$-fold.
In contrast, $\widetilde{\pi}$ caters to all possible starting robot locations. So, the incurred time to derive $\widetilde{\pi}$ is unaffected.
\begin{figure*}
\hspace{-1mm}
\begin{tabular}{cccc}
\hspace{-2mm}\epsfig{figure=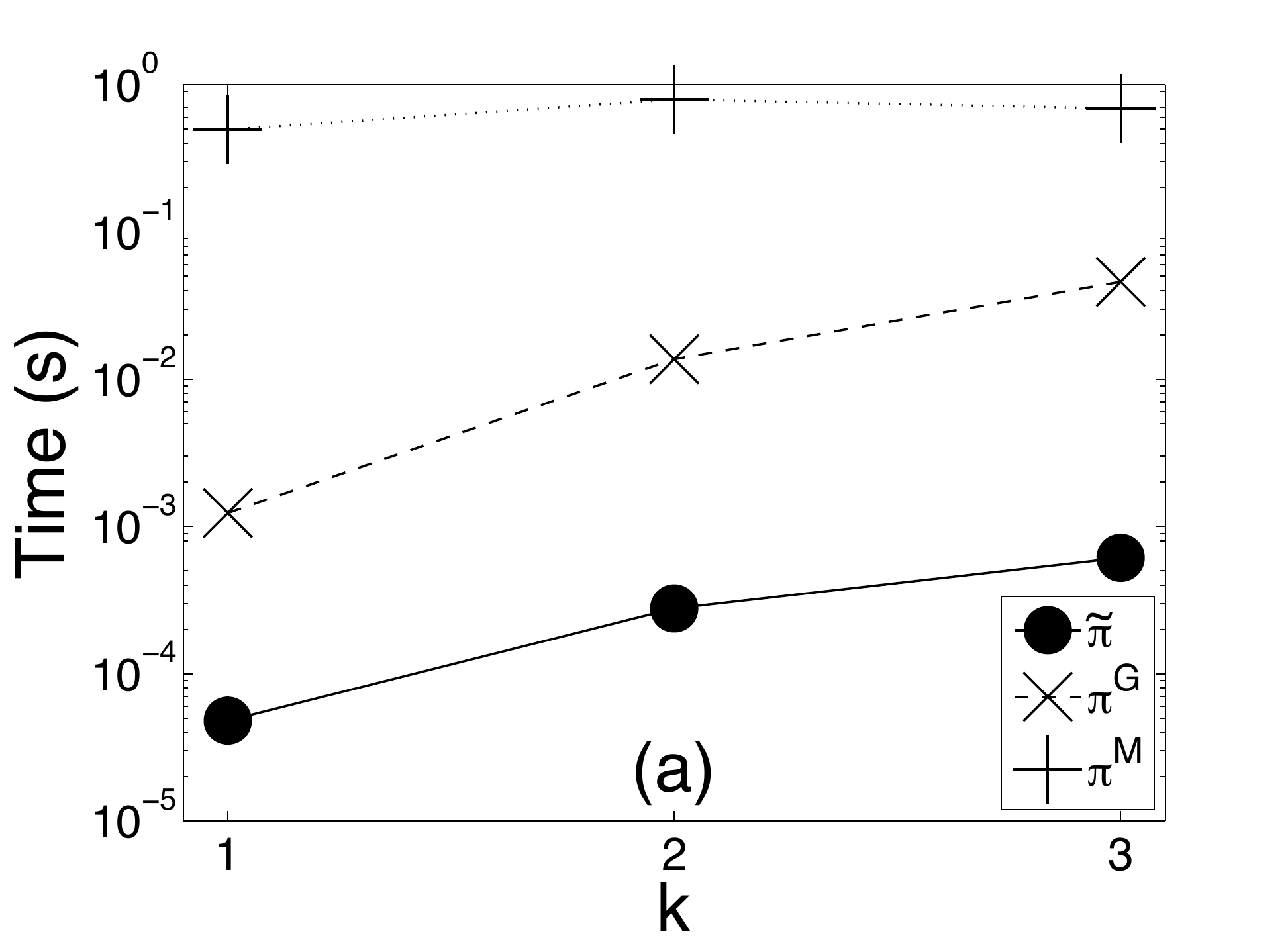,width=4.8cm} & \hspace{-7mm}\epsfig{figure=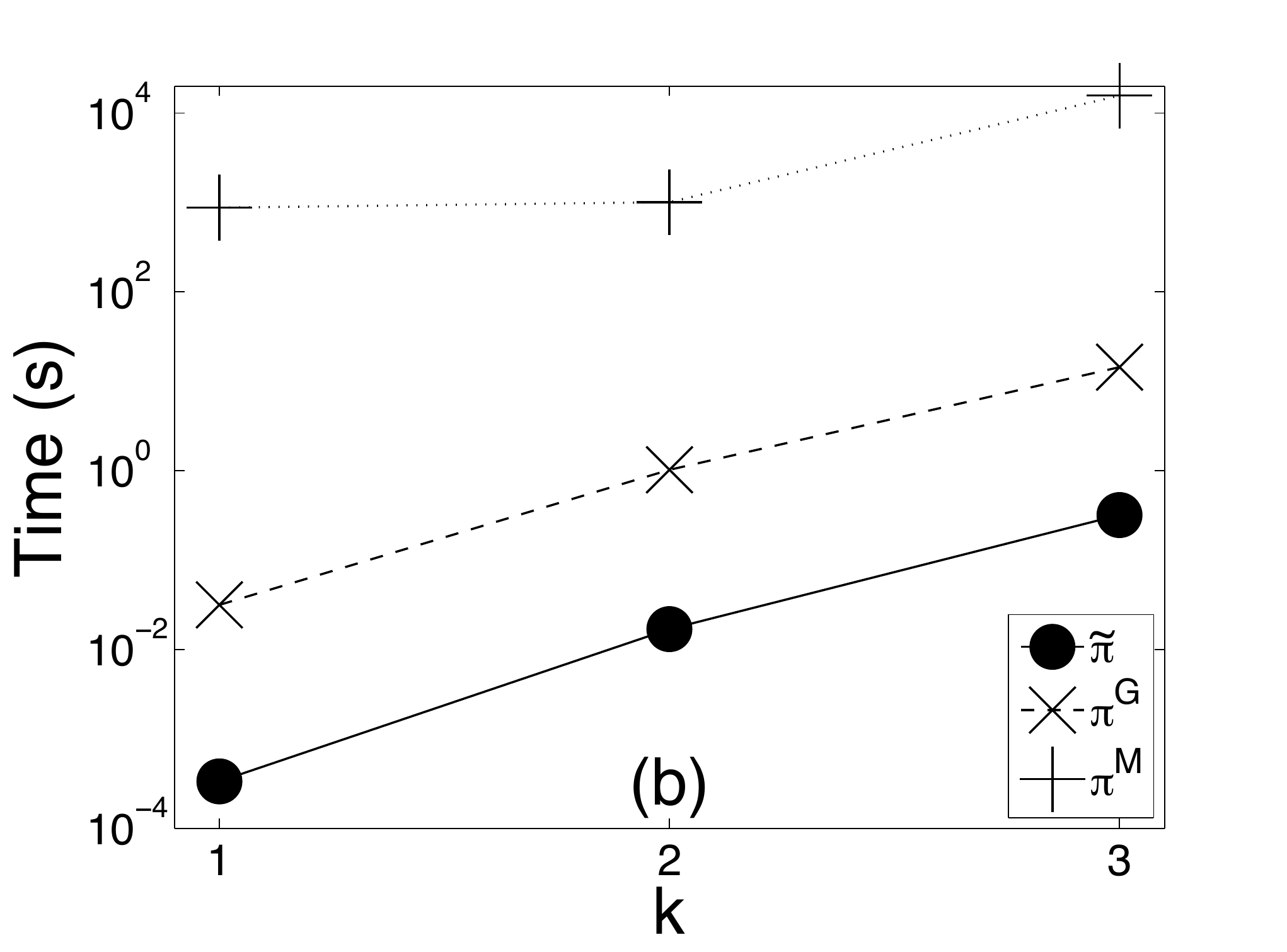,width=4.8cm} & \hspace{-7mm}\epsfig{figure=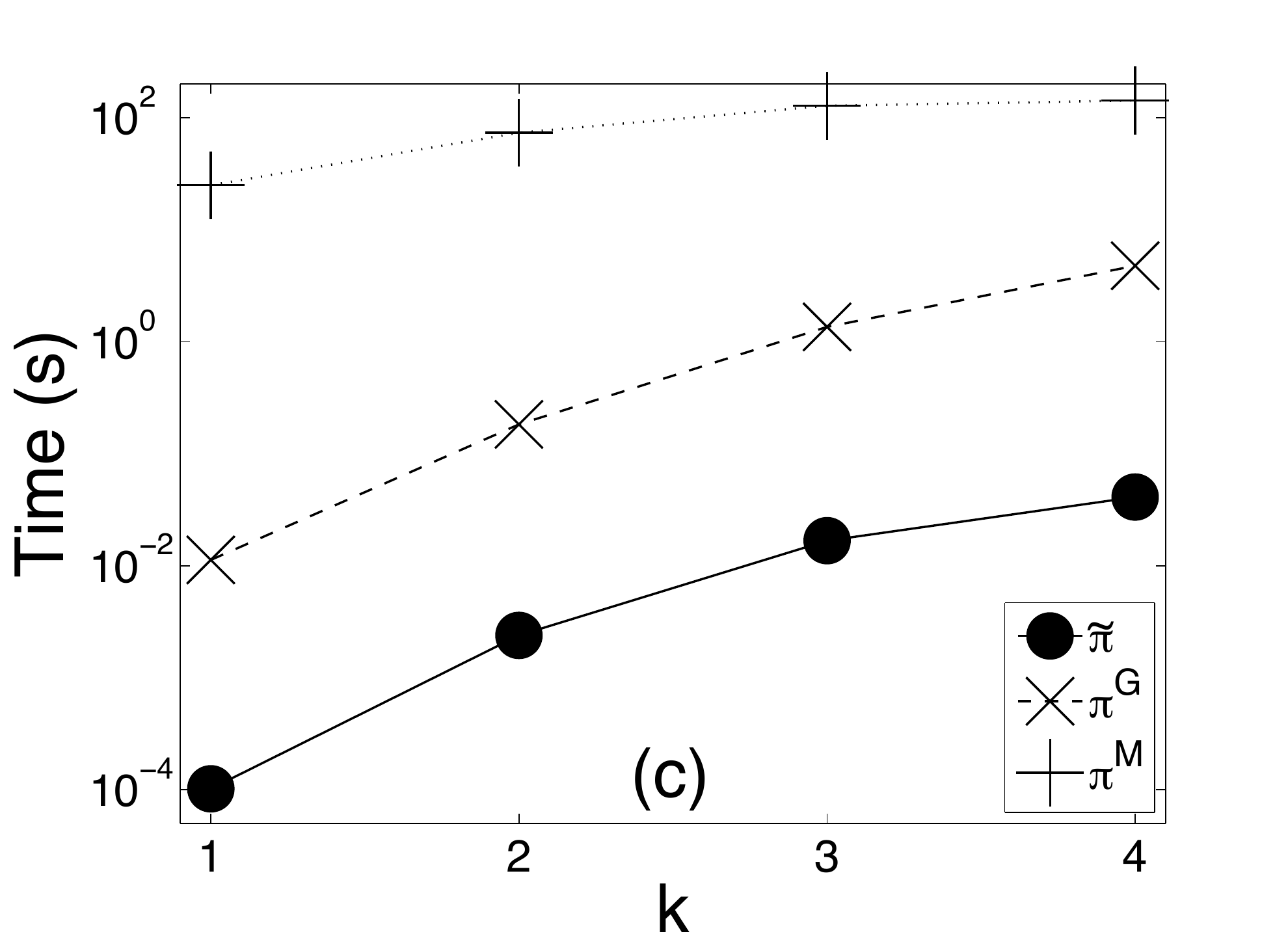,width=4.8cm} & \hspace{-7mm}\epsfig{figure=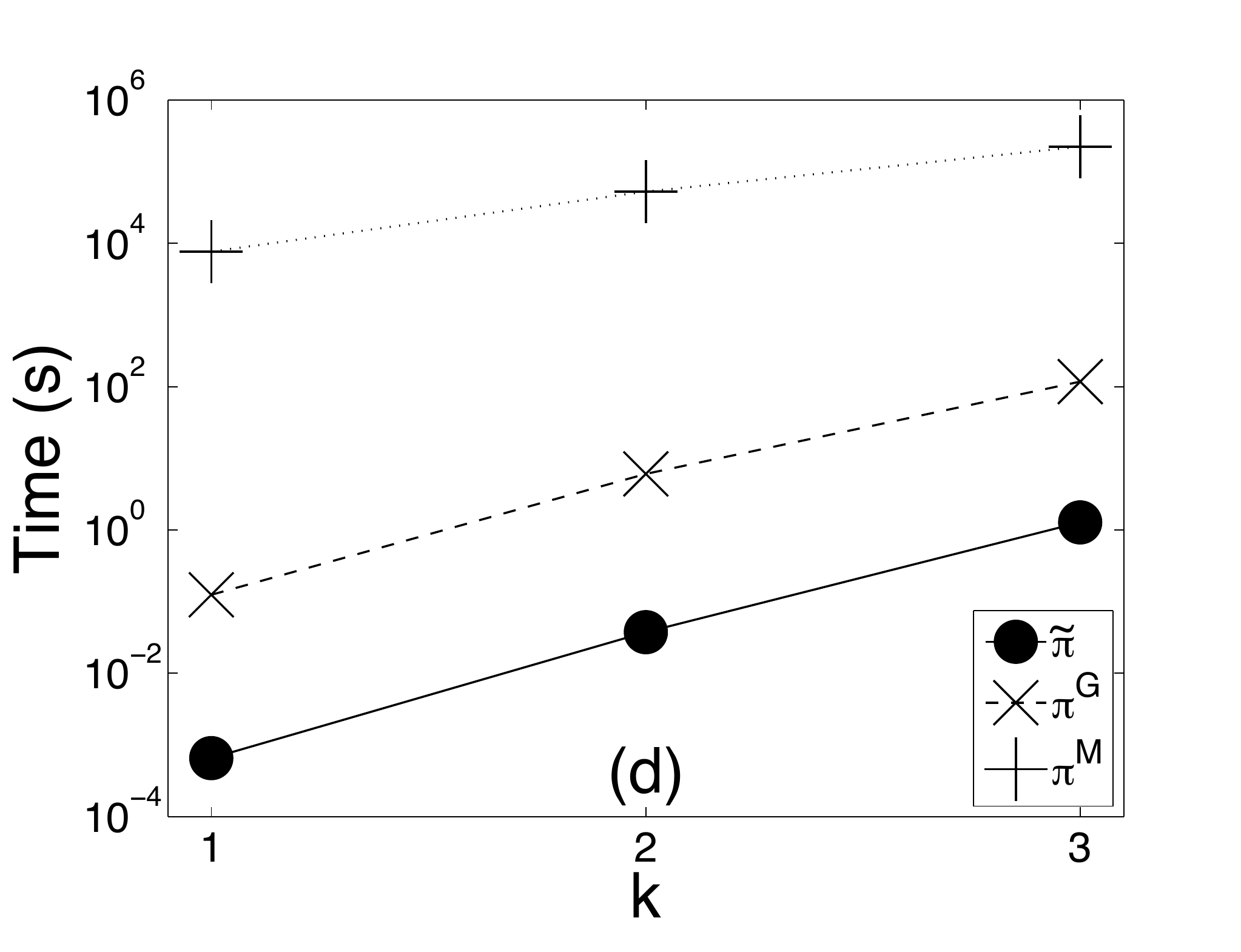,width=4.8cm}
\vspace{-5mm}
\end{tabular}
\caption{Graph of time taken to derive policy vs. number $k$ of robots for temperature field $4$ discretized into (a) $5 \times 30$ and (b) $13 \times 75$ grids and plankton density field discretized into (c) $8 \times 45$ and (d) $16 \times 89$ grids.}
\vspace{-3mm}
\label{fig:9.4}
\end{figure*}
These observations show a considerable computational gain of $\widetilde{\pi}$ over $\pi^G$ and $\pi^M$, which supports our time complexity analysis and comparison (Section~\ref{sect:mpp}).
So, our Markov-based path planner is more time-efficient for \emph{in situ}, real-time, high-resolution active sampling.
%
\section{Conclusion}
\begin{table}
\vspace{-3mm}
\caption{Comparison of ENT($\pi$) (left) and ERR($\pi$)  ($\times 10^{-3}$) (right) performance for plankton density field that is discretized into $8 \times 45$ grid.}
\label{tab:cl}
\begin{tiny}
\begin{tabular}{cc}
\hspace{-11mm}
\begin{tabular}{|c|c|c|c|c|}
\hline
ENT($\pi$) & \multicolumn{4}{|c|}{Number $k$ of robots}\\
\hline
Policy & 1 & 2 & 3 & 4 \\
\hline
$\widetilde{\pi}$ & -359 & -322 & -196 & -121 \\
$\pi^G$ & -359 & -322 & -196 & -121 \\
$\pi^{M}$ & -230 & -186 & -70 & -11 \\
\hline
\end{tabular}
&\hspace{-4mm}
\begin{tabular}{|c|c|c|c|c|}
\hline
ERR($\pi$) & \multicolumn{4}{|c|}{Number $k$ of robots}\\
\hline
Policy & 1 & 2 & 3 & 4 \\
\hline
$\widetilde{\pi}$ & 5.6124 & 2.2164 & 0.0544 & 0.0066 \\
$\pi^G$ & 5.6124 & 2.2164 & 0.0544 & 0.0066 \\
$\pi^{M}$ & 4.5371 & 0.5613 & 0.0472 & 0.0324 \\
\hline
\end{tabular}
\end{tabular}
\vspace{-4mm}
\end{tiny}
\end{table}
\begin{table}
\caption{Comparison of ENT($\pi$) (left) and ERR($\pi$)  ($\times 10^{-3}$) (right) performance for plankton density field that is discretized into $16 \times 89$ grid.}
\label{tab:ch}
\begin{tiny}
\begin{tabular}{cc}
\hspace{-3mm}
\begin{tabular}{|c|c|c|c|c|}
\hline
ENT($\pi$) & \multicolumn{3}{|c|}{Number $k$ of robots}\\
\hline
Policy & 1 & 2 & 3 \\
\hline
$\widetilde{\pi}$ & -4278 & -3949 & -3681 \\
$\pi^G$ & -4238 & -3964 & -3686 \\
$\pi^{M}$ & -4171 & -3840 & -3501 \\
\hline
\end{tabular}
&\hspace{-4mm}
\begin{tabular}{|c|c|c|c|c|}
\hline
ERR($\pi$) & \multicolumn{3}{|c|}{Number $k$ of robots}\\
\hline
Policy & 1 & 2 & 3 \\
\hline
$\widetilde{\pi}$ & 3.4328 & 0.0970 & 0.0546 \\
$\pi^G$ & 1.5648 & 0.1073 & 0.0643 \\
$\pi^{M}$ & 0.8186 & 0.0859 & 0.0348 \\
\hline
\end{tabular}
\end{tabular}
\end{tiny}
\vspace{-3mm}
\end{table}
This paper describes an efficient Markov-based information-theoretic path planner for active sampling of GP-based environmental fields.
We have provided theoretical guarantees on the active sampling performance of our Markov-based policy $\widetilde{\pi}$ for the transect sampling task, 
from which ideal environmental field conditions (i.e., small horizontal spatial correlation and noisy, less intense fields) and sampling task settings (i.e., large grid discretization width and short planning horizon) can be established to limit its performance degradation. 
Empirically, we have shown that $\widetilde{\pi}$ can generally achieve active sampling performance comparable to that of the widely-used non-Markovian greedy policies $\pi^G$ and $\pi^M$ under less favorable realistic field conditions (i.e., low noise-to-signal ratio) and task settings (i.e., small grid discretization width and long planning horizon) while enjoying huge computational gain over them. 
In particular, we have empirically observed that (a) small horizontal and large vertical correlations strongly favor $\widetilde{\pi}$; (b) though large horizontal and small vertical correlations do not favor $\widetilde{\pi}$, this problem can be mitigated by increasing the number of robots.
In fact, deploying a large robot team often produces superior active sampling performance of $\widetilde{\pi}$ over $\pi^M$ in our experiments, not forgetting the computational gain of $> 4$ orders of magnitude.
Our Markov-based planner can be used to efficiently achieve more general exploration tasks (e.g., boundary tracking and those in \cite{LowAAMAS08, LowICAPS09}), but the guarantees provided here may not apply.
For our future work, we will ``relax'' the Markov assumption by utilizing a longer (but not entire) history of observations in path planning. This can potentially improve the active sampling performance in fields of moderate to large horizontal correlation but does not incur as much time as that of non-Markovian policies.
\bibliographystyle{abbrv}
\bibliography{adaptivesampling}


\appendix
\section{Proofs}
\label{sect:proofs}

\subsection{Proof Sketch of Theorem~\ref{thm:1}}
For each vector ${x}_i$ of current robot locations, the time needed to evaluate the posterior entropy $\mathbb{H}[{Z}_{\tau({x}_{i}, {a}_{i})}| Z_{{x}_{i}}]$ (i.e., using Cholesky factorization) over all possible actions ${a}_i\in\set{A}({x}_i)$ is $|\set{A}| \times\set{O}(k^4) = \set{O}(|\set{A}| k^4)$. 
Doing this over all possible vectors of current robot locations in each column thus incurs $|\set{A}|\times\set{O}(|\set{A}| k^4)=\set{O}(|\set{A}|^2 k^4)$ time since the vector space of current robot locations in each column is of the same size as that of the joint action space $|\set{A}|$. 
We do not have to compute these posterior entropies again for every column because the entropies evaluated for any one column replicate across different columns.
This computational saving is due to the Markov assumption and the problem structure of the transect sampling task.
Propagating the optimal values from stages $t$ to $0$ takes $\set{O}(|\set{A}|^2 t)$ time. 
Hence, solving the Markov-based path planning problem (\ref{eq:9.2}) or deriving the Markov-based policy $\widetilde{\pi}$ (\ref{eq:9.2op}) takes $\set{O}(|\set{A}|^2 (t+k^4))$ time for the transect sampling task.

\subsection{Proof of Lemma~\ref{lem:9.7}}
\label{sect:app16}
Let $\displaystyle\Sigma_{{x}_{0:i-1} {x}_{0:i-1}\mid {x}_{i}} \defeq C+E$ where $C$ is defined to be a matrix with diagonal components $\sigma^2_{{x_k}} = \sigma^2_s + \sigma^2_n$ for $k = 0,\ldots,i-1$ and off-diagonal components $0$, and $E$ is defined to be a matrix with diagonal components $-(\sigma_{{x_k}{x_i}})^2/\sigma^2_{{x_i}} = -(\sigma_{{x_k}{x_i}})^2/(\sigma^2_s + \sigma^2_n)$ for $k = 0,\ldots,i-1$ and the same off-diagonal components as $\Sigma_{{x}_{0:i-1} {x}_{0:i-1}\mid {x}_{i}}$ (i.e., $\sigma_{{x_{j}}{x_k}\mid x_i} = \sigma_{{x_{j}}{x_k}} - \sigma_{{x_{j}}{x_i}}\sigma_{{x_{i}}{x_k}}/\sigma^2_{{x_i}}$ for $j,k = 0,\ldots,i-1$, $j\neq k$). Then, 
\begin{equation}
||C^{-1}||_2 = ||(\sigma^2_s + \sigma^2_n)^{-1}{I}||_2 = \frac{1}{\sigma^2_s + \sigma^2_n} \ .
\label{eq:app16.1}
\end{equation}
The last equality follows from $\sigma^2_s + \sigma^2_n$ being the smallest eigenvalue of $C$. So, $1/(\sigma^2_s + \sigma^2_n)$ is the largest eigenvalue of $C^{-1}$, which is equal to $||C^{-1}||_2$. 

Note that the minimum distance between any pair of location components of ${x}_{0:i-1}$ cannot be less than $\omega_1$. So, it can be observed that any component of $E$ cannot have an absolute value more than $\sigma^2_s \xi$.
Therefore,
\begin{equation}
||E||_2 \leq \displaystyle i \sigma^2_s \xi \ ,
\label{eq:app16.2}
\end{equation}
which follows from a property of the matrix $2$-norm that 
$||E||_2$ cannot be more than the largest absolute component of $E$ multiplied by $i$ \cite{Golub96}.

Note that the minimum distance between locations $x_i$ and $x_{i+1}$ as well as between location $x_i$ and any location component of ${x}_{0:i-1}$ cannot be less than $\omega_1$.
So, it can be observed that any component of $\Sigma_{x_{i+1} {x}_{0:i-1}\mid {x}_{i}}$ cannot have an absolute value more than $\sigma^2_s \xi^2$. Therefore,
\begin{equation}
|\sigma_{Z_{x_{i+1}}Z_{x_k}\mid x_i}| \leq \sigma^2_s \xi^2
\label{eq:app16.3}
\end{equation}
for $k = 0,\ldots,i-1$.

Now,
\begin{equation}
\begin{array}{l}
\Sigma_{x_{i+1} {x}_{0:i-1}\mid {x}_{i}} (C+E)^{-1} \Sigma_{{x}_{0:i-1} x_{i+1}\mid {x}_{i}} \ - \vspace{1mm}\\
\Sigma_{x_{i+1} {x}_{0:i-1}\mid {x}_{i}} C^{-1} \Sigma_{{x}_{0:i-1} x_{i+1}\mid {x}_{i}}\vspace{1mm}\\
= \Sigma_{x_{i+1} {x}_{0:i-1}\mid {x}_{i}} \{(C+E)^{-1} - C^{-1}\} \Sigma_{{x}_{0:i-1} x_{i+1}\mid {x}_{i}}\vspace{1mm}\\
\leq ||\Sigma_{x_{i+1} {x}_{0:i-1}\mid {x}_{i}}||^2_2 \ ||(C+E)^{-1} - C^{-1}||_2\vspace{1mm}\\
\leq \displaystyle \sum^{i-1}_{k=0} |\sigma_{Z_{x_{i+1}}Z_{x_k}\mid x_i}|^2 \frac{||C^{-1}||_2 \ ||E||_2}{\frac{1}{||C^{-1}||_2} - ||E||_2}\vspace{1mm}\\
= \displaystyle i (\sigma^2_s)^2 \xi^4 \frac{||C^{-1}||_2 \ ||E||_2}{\frac{1}{||C^{-1}||_2} - ||E||_2} \ .
\end{array}
\label{eq:app16.4}
\end{equation}
The first inequality is due to Cauchy-Schwarz inequality and submultiplicativity of the matrix norm \cite{Stewart90}.
The second inequality follows from an important result in the perturbation theory of matrix inverses (in particular, Theorem III.2.5 in \cite{Stewart90}). It requires the assumption of $||C^{-1} \ E||_2 < 1$. This assumption can be satisfied by $||C^{-1}||_2 \ ||E||_2 < 1$ because $||C^{-1} \ E||_2\leq ||C^{-1}||_2 \ ||E||_2$.
By (\ref{eq:app16.1}) and (\ref{eq:app16.2}), $||C^{-1}||_2 \ ||E||_2 < 1$ translates to $\xi < \rho/i$.
The last equality is due to (\ref{eq:app16.3}).

From (\ref{eq:app16.4}),
\begin{equation}
\begin{array}{l}
\Sigma_{x_{i+1} {x}_{0:i-1}\mid {x}_{i}} (C+E)^{-1} \Sigma_{{x}_{0:i-1} x_{i+1}\mid {x}_{i}} \vspace{1mm}\\
\leq \displaystyle \Sigma_{x_{i+1} {x}_{0:i-1}\mid {x}_{i}} C^{-1} \Sigma_{{x}_{0:i-1} x_{i+1}\mid {x}_{i}}+ i (\sigma^2_s)^2 \xi^4 \frac{||C^{-1}||_2 \ ||E||_2}{\frac{1}{||C^{-1}||_2} - ||E||_2}\vspace{1mm}\\
\leq \displaystyle i (\sigma^2_s)^2 \xi^4 \ ||C^{-1}||_2 \left( 1+ \frac{||E||_2}{\frac{1}{||C^{-1}||_2} - ||E||_2}\right) \vspace{1mm}\\
= \displaystyle \frac{i (\sigma^2_s)^2 \xi^4}{\frac{1}{||C^{-1}||_2} - ||E||_2}\vspace{1mm}\\
\leq\displaystyle\frac{i (\sigma^2_s)^2 \xi^4}{\sigma^2_s + \sigma^2_n  - i \sigma^2_s \xi}\vspace{1mm}\\
=\displaystyle\frac{\sigma^2_s \xi^4}{\frac{\rho}{i}  - \xi}
\end{array}
\label{eq:app16.6}
\end{equation}
The second inequality is due to 
$$
\displaystyle\Sigma_{x_{i+1} {x}_{0:i-1}\mid {x}_{i}} C^{-1} \Sigma_{{x}_{0:i-1} x_{i+1}\mid {x}_{i}} \leq i (\sigma^2_s)^2 \xi^4 \ ||C^{-1}||_2 \ ,
$$
which follows from Cauchy-Schwarz inequality and (\ref{eq:app16.3}).
The third inequality follows from  (\ref{eq:app16.1}) and  (\ref{eq:app16.2}). 

We will need the following property of posterior variance that is similar to (\ref{eq:7}):
\begin{equation}
\sigma^2_{{{x}_{i+1}}\mid {x}_{0:i}} = \sigma^2_{{{x}_{i+1}}\mid {x}_{i}}-\Sigma_{x_{i+1} {x}_{0:i-1}\mid {x}_{i}} \Sigma^{-1}_{{x}_{0:i-1} {x}_{0:i-1}\mid {x}_{i}} \Sigma_{{x}_{0:i-1} x_{i+1}\mid {x}_{i}}
\label{eq:9.7}
\end{equation}
where $\Sigma_{x_{i+1} {x}_{0:i-1}\mid {x}_{i}}$ is a posterior covariance vector with components $\sigma_{{x_{i+1}}{x_k}\mid {x}_{i}}$ for $k=0,\ldots,i-1$,
$\Sigma_{{x}_{0:i-1} x_{i+1}\mid {x}_{i}}$ is the transpose of
$\Sigma_{x_{i+1} {x}_{0:i-1}\mid {x}_{i}}$,
and $\Sigma_{{x}_{0:i-1} {x}_{0:i-1}\mid {x}_{i}}$ is a posterior covariance matrix with components $\sigma_{{x_j}{x_k}\mid {x}_{i}}$ for $j,k = 0,\ldots,i-1$.

By (\ref{eq:app16.6}) and (\ref{eq:9.7}), 
$$
\begin{array}{l}
\sigma^2_{{{x}_{i+1}}\mid {x}_{i}}-\sigma^2_{{{x}_{i+1}}\mid {x}_{0:i}}\vspace{1mm}\\
= \Sigma_{x_{i+1} {x}_{0:i-1}\mid {x}_{i}} \Sigma^{-1}_{{x}_{0:i-1} {x}_{0:i-1}\mid {x}_{i}} \Sigma_{{x}_{0:i-1} x_{i+1}\mid {x}_{i}}\vspace{1mm}\\
\leq \displaystyle\frac{\sigma^2_s \xi^4}{\frac{\rho}{i}  - \xi} \ .
\end{array}
$$
\subsection{Proof of Theorem~\ref{thm:9.9}}
\label{sect:app17}
\emph{Proof by induction} on $i$ that ${V}^{\pi^{\ast}}_{i}({x}_{0:i}) \leq \widetilde{V}_{i}({x}_{i}) \leq {V}^{\pi^{\ast}}_{i}({x}_{0:i}) \ + \sum^t_{s=i} \Delta(s)$ for $i = t,\ldots,0$.\\
\\
\noindent
\emph{Base case} ($i = t$): By Lemma~\ref{lem:9.8},
\begin{equation}
\begin{array}{rl}
&\mathbb{H}[{Z}_{{x}_{t+1}}| {Z}_{{x}_{0:t}}] \leq \mathbb{H}[{Z}_{{x}_{t+1}}| {Z}_{{x}_{t}}]\vspace{1mm}\\  
&\leq \mathbb{H}[{Z}_{{x}_{t+1}}| {Z}_{{x}_{0:t}}]+\Delta(t) \ \ \ \mbox{for any} \ {x}_{t+1}\vspace{1mm}\\
\Rightarrow & \displaystyle\max_{{a}_{t} \in \set{A}({x}_{t})} \mathbb{H}[{Z}_{{x}_{t+1}}| {Z}_{{x}_{0:t}}] \leq \displaystyle\max_{{a}_{t} \in \set{A}({x}_{t})}\mathbb{H}[{Z}_{{x}_{t+1}}| {Z}_{{x}_{t}}]\vspace{1mm}\\  
& \leq \displaystyle\max_{{a}_{t} \in \set{A}({x}_{t})}\mathbb{H}[{Z}_{{x}_{t+1}}| {Z}_{{x}_{0:t}}]+\Delta(t) \vspace{1mm}\\
\Rightarrow & {V}^{\pi^{\ast}}_{t}({x}_{0:t}) \leq \widetilde{V}_{t}({x}_{t}) \leq {V}^{\pi^{\ast}}_{t}({x}_{0:t}) + \Delta(t) \ .
\end{array}
\label{eq:app17.1}
\end{equation}
Hence, the base case is true.\\
\\
\noindent
\emph{Inductive case}: Suppose that 
\begin{equation}
{V}^{\pi^{\ast}}_{i+1}({x}_{0:i+1}) \leq \widetilde{V}_{i+1}({x}_{i+1}) \leq {V}^{\pi^{\ast}}_{i+1}({x}_{0:i+1}) + \sum^t_{s=i+1} \Delta(s)
\label{eq:app17.2}
\end{equation}
is true. We have to prove that 
${V}^{\pi^{\ast}}_{i}({x}_{0:i}) \leq \widetilde{V}_{i}({x}_{i}) \leq {V}^{\pi^{\ast}}_{i}({x}_{0:i}) \ + \sum^t_{s=i} \Delta(s)$ is true.

We will first show that $\widetilde{V}_{i}({x}_{i}) \leq {V}^{\pi^{\ast}}_{i}({x}_{0:i}) + \sum^t_{s=i} \Delta(s)$. By Lemma~\ref{lem:9.8},
$$
\begin{array}{rl}
&\displaystyle\mathbb{H}[{Z}_{{x}_{i+1}}| {Z}_{{x}_{i}}]  \leq \mathbb{H}[{Z}_{{x}_{i+1}}| {Z}_{{x}_{0:i}}]+\Delta(i) \ \ \ \mbox{for any} \ {x}_{i+1}\vspace{1mm}\\
\Rightarrow &\displaystyle\mathbb{H}[{Z}_{{x}_{i+1}}| {Z}_{{x}_{i}}] + \widetilde{V}_{i+1}({x}_{i+1}) \leq \mathbb{H}[{Z}_{{x}_{i+1}}| {Z}_{{x}_{0:i}}] \ +\vspace{1mm}\\
& {V}^{\pi^{\ast}}_{i+1}({x}_{0:i+1}) + \sum^t_{s=i} \Delta(s) \ \mbox{by} \ (\mbox{\ref{eq:app17.2}}) \ \mbox{for any} \ {x}_{i+1}\vspace{1mm}\\
\Rightarrow &\displaystyle\max_{{a}_{i} \in \set{A}({x}_{i})}\mathbb{H}[{Z}_{{x}_{i+1}}| {Z}_{{x}_{i}}] + \widetilde{V}_{i+1}({x}_{i+1})\\ 
& \leq \displaystyle\max_{{a}_{i} \in \set{A}({x}_{i})}\mathbb{H}[{Z}_{{x}_{i+1}}| {Z}_{{x}_{0:i}}] +{V}^{\pi^{\ast}}_{i+1}({x}_{0:i+1}) + \sum^t_{s=i} \Delta(s)\vspace{1mm}\\
\Rightarrow &\displaystyle\widetilde{V}_{i}({x}_{i}) \leq {V}^{\pi^{\ast}}_{i}({x}_{0:i}) + \sum^t_{s=i} \Delta(s) \ .
\end{array}
$$
We will now prove that ${V}^{\pi^{\ast}}_{i}({x}_{0:i}) \leq \widetilde{V}_{i}({x}_{i})$. By Lemma~\ref{lem:9.8},
$$
\begin{array}{rl}
&\mathbb{H}[{Z}_{{x}_{i+1}}| {Z}_{{x}_{0:i}}] \leq \mathbb{H}[{Z}_{{x}_{i+1}}| {Z}_{{x}_{i}}] \ \ \ \mbox{for any} \ {x}_{i+1}\vspace{1mm}\\ 
\Rightarrow &\displaystyle\mathbb{H}[{Z}_{{x}_{i+1}}| {Z}_{{x}_{0:i}}] + {V}^{\pi^{\ast}}_{i+1}({x}_{0:i+1})\vspace{1mm}\\
& \leq \mathbb{H}[{Z}_{{x}_{i+1}}| {Z}_{{x}_{i}}] + \widetilde{V}_{i+1}({x}_{i+1}) \ \mbox{by} \ (\mbox{\ref{eq:app17.2}}) \ \mbox{for any} \ {x}_{i+1}\vspace{1mm}\\
\Rightarrow &\displaystyle\max_{{a}_{i} \in \set{A}({x}_{i})}\mathbb{H}[{Z}_{{x}_{i+1}}| {Z}_{{x}_{0:i}}] + {V}^{\pi^{\ast}}_{i+1}({x}_{0:i+1})\vspace{1mm}\\
& \leq \max_{{a}_{i} \in \set{A}({x}_{i})}\mathbb{H}[{Z}_{{x}_{i+1}}| {Z}_{{x}_{i}}] + \widetilde{V}_{i+1}({x}_{i+1})\vspace{1mm}\\
\Rightarrow &\displaystyle{V}^{\pi^{\ast}}_{i}({x}_{0:i}) \leq \widetilde{V}_{i}({x}_{i}) \ .
\end{array}
$$
Hence, the inductive case is true.
\subsection{Proof of Theorem~\ref{thm:9.10}}
\label{sect:app18}
The following lemma is needed for this proof:
\begin{lemma}
$\widetilde{V}_{i}({x}_{i}) \leq {V}^{\widetilde{\pi}}_{i}({x}_{0:i}) + \sum^t_{s=i} \Delta(s)$ for $i = 0,\ldots,t$.
\label{lem:app18.1}
\end{lemma}
The proof of the above lemma is provided in Appendix~\ref{sect:app19}.

\emph{Proof by induction} on $i$ that ${V}^{\pi^{\ast}}_{i}({x}_{0:i}) \leq {V}^{\widetilde{\pi}}_{i}({x}_{0:i}) + \sum^t_{s=i} \Delta(s)$ for $i = t,\ldots,0$.\\
\\
\noindent
\emph{Base case} ($i = t$):
$$
{V}^{\pi^{\ast}}_{t}({x}_{0:t}) \leq \widetilde{V}_{t}({x}_{t}) \leq {V}^{\widetilde{\pi}}_{t}({x}_{0:t}) +\Delta(t) \ .
$$
The first inequality is due to Theorem~\ref{thm:9.9}. The second inequality follows from Lemma~\ref{lem:app18.1}. Hence, the base case is true.\\
\\
\noindent
\emph{Inductive case}: Suppose that 
\begin{equation}
{V}^{\pi^{\ast}}_{i+1}({x}_{0:i+1}) \leq {V}^{\widetilde{\pi}}_{i+1}({x}_{0:i+1}) + \sum^t_{s=i+1} \Delta(s)
\label{eq:app18.2}
\end{equation}
is true. We have to prove that 
${V}^{\pi^{\ast}}_{i}({x}_{0:i}) \leq {V}^{\widetilde{\pi}}_{i}({x}_{0:i}) + \sum^t_{s=i} \Delta(s)$ is true.
$$
\begin{array}{l}
{V}^{\pi^{\ast}}_{i}({x}_{0:i}) \leq \widetilde{V}_{i}({x}_{i})\vspace{1mm}\\
=\displaystyle\mathbb{H}[{Z}_{\tau(x_i,\widetilde{\pi}_i(x_i))}|{Z}_{{x}_{i}}] + \widetilde{V}_{i+1}(\tau(x_i,\widetilde{\pi}_i(x_i)))\vspace{1mm}\\
\leq \displaystyle\mathbb{H}[{Z}_{\tau(x_i,\widetilde{\pi}_i({x}_{i}))}|{Z}_{{x}_{0:i}}] + \Delta(i) + \widetilde{V}_{i+1}(\tau(x_i,\widetilde{\pi}_i(x_i)))\vspace{1mm}\\
\leq \displaystyle\mathbb{H}[{Z}_{\tau(x_i,\widetilde{\pi}_i({x}_{i}))}|{Z}_{{x}_{0:i}}] + \Delta(i) +{V}^{\widetilde{\pi}}_{i+1}(
\ ({x}_{0:i}, \tau({x}_{i}, \widetilde{\pi}_i({x}_{i})))\ 
) \ + \vspace{1mm}\\
\ \ \ \sum^t_{s=i+1} \Delta(s)\vspace{1mm}\\
= \displaystyle{V}^{\widetilde{\pi}}_{i}({x}_{0:i}) + \sum^t_{s=i} \Delta(s) \ .
\end{array}
$$
The first inequality is due to Theorem~\ref{thm:9.9}. 
The first equality follows from (\ref{eq:9.2}).
The second inequality follows from Lemma~\ref{lem:9.8}.
The third inequality is due to Lemma~\ref{lem:app18.1}. 
The last equality follows from (\ref{eq:9.9a}).
Hence, the inductive case is true.
\subsection{Proof Sketch of Lemma~\ref{lem:9.11}}
Define $x^{[m]}_i$ to be the $m$-th component of vector $x_i$ of robot locations for $m = 1,\ldots,k$. Let $x^{[1:m]}_i$ denote a vector comprising the first $m$ components of $x_i$ (i.e., concatenation of $x^{[1]}_i,\ldots,x^{[m]}_i$).
\begin{equation}
\begin{array}{l}
\mathbb{I}[{Z}_{{x}_{i+1}};Z_{{x}_{0:i-1}}|Z_{{x}_{i}}]\vspace{1mm}\\
= \mathbb{H}[{Z}_{{x}_{i+1}}| Z_{{x}_{i}}] - \mathbb{H}[{Z}_{{x}_{i+1}}| Z_{{x}_{0:i}}]\vspace{1mm}\\
= \displaystyle\sum^k_{m=1} \left( \mathbb{H}[{Z}_{{x}^{[m]}_{i+1}}|Z_{(x_i,{x}^{[1:m-1]}_{i+1})}] - \mathbb{H}[{Z}_{{x}^{[m]}_{i+1}}|Z_{({x}_{0:i}, {x}^{[1:m-1]}_{i+1})}] \right)\vspace{1mm}\\
= \displaystyle \frac{1}{2} \sum^k_{m=1} \left(\log\frac{\sigma^2_{{{x}^{[m]}_{i+1}}| (x_i,{x}^{[1:m-1]}_{i+1})}}{\sigma^2_{{{x}^{[m]}_{i+1}}|({x}_{0:i},{x}^{[1:m-1]}_{i+1})}}\right)\vspace{1mm}\\
= \displaystyle \frac{1}{2} \sum^k_{m=1} \log \left( 1 - \frac{\sigma^2_{{{x}^{[m]}_{i+1}}| (x_i,{x}^{[1:m-1]}_{i+1})}-\sigma^2_{{{x}^{[m]}_{i+1}}|({x}_{0:i},{x}^{[1:m-1]}_{i+1})}}{\sigma^2_{{{x}^{[m]}_{i+1}}| (x_i,{x}^{[1:m-1]}_{i+1})}} \right)^{-1}\vspace{1mm}\\
\geq 0 \ .
\end{array}
\label{eq:app20.1}
\end{equation}
The second equality follows from the chain rule for entropy.

Similar to Lemma~\ref{lem:9.7}, the following result bounds the variance reduction term $$\sigma^2_{{{x}^{[m]}_{i+1}}| (x_i,{x}^{[1:m-1]}_{i+1})}-\sigma^2_{{{x}^{[m]}_{i+1}}|({x}_{0:i},{x}^{[1:m-1]}_{i+1})}$$ in (\ref{eq:app20.1}):
\begin{lemma} 
If $\displaystyle \xi < \min(\frac{\rho}{ik},\frac{\rho}{4k})$ and (\ref{eq:9.10}) is satisfied,
$$\displaystyle 0 \leq\sigma^2_{{{x}^{[m]}_{i+1}}| (x_i,{x}^{[1:m-1]}_{i+1})}-\sigma^2_{{{x}^{[m]}_{i+1}}|({x}_{0:i},{x}^{[1:m-1]}_{i+1})}\leq\hspace{-0mm}\frac{\sigma^2_s \xi^4}{\frac{\rho}{ik}  - \xi} \ .$$
\label{lem:app20.2}
\end{lemma}
The proof of the above result is largely similar to that of Lemma~\ref{lem:9.7} (Appendix~\ref{sect:app16}), and is therefore omitted here.

The bounds on $\mathbb{I}[{Z}_{{x}_{i+1}};Z_{{x}_{0:i-1}}|Z_{{x}_{i}}]$ follow immediately from (\ref{eq:app20.1}), Lemma~\ref{lem:app20.2}, and the following lower bound on $\sigma^2_{{{x}^{[m]}_{i+1}}| (x_i,{x}^{[1:m-1]}_{i+1})}$:
$$
\begin{array}{l}
\sigma^2_{{{x}^{[m]}_{i+1}}| (x_i,{x}^{[1:m-1]}_{i+1})} \vspace{1mm}\\
= \sigma^2_{{{x}^{[m]}_{i+1}}} \ -\vspace{1mm}\\ 
\ \ \ \Sigma_{{{x}^{[m]}_{i+1}} (x_i,{x}^{[1:m-1]}_{i+1})} \Sigma^{-1}_{(x_i,{x}^{[1:m-1]}_{i+1}) (x_i,{x}^{[1:m-1]}_{i+1})} \Sigma_{(x_i,{x}^{[1:m-1]}_{i+1}) {{x}^{[m]}_{i+1}}}\vspace{1mm}\\
\geq \displaystyle\sigma^2_s + \sigma^2_n - \frac{\sigma^2_s\xi^2}{\frac{\rho}{2k-1}-\xi}\vspace{1mm}\\
\geq \sigma^2_s + \sigma^2_n - \frac{4k}{\rho} \sigma^2_s \xi^2 \ .
\end{array}
$$
The equality is due to (\ref{eq:7}).
The first inequality is due to Cauchy-Schwarz inequality, submultiplicativity of the matrix norm \cite{Stewart90}, and a result in the perturbation theory of matrix inverses (in particular, Theorem III.2.5 in \cite{Stewart90}).
The second inequality follows from the given satisfied condition $\displaystyle \xi < \frac{\rho}{4k}$.
\subsection{Proof of Lemma~\ref{lem:app18.1}}
\label{sect:app19}
\emph{Proof by induction} on $i$ that $\widetilde{V}_{i}({x}_{i}) \leq {V}^{\widetilde{\pi}}_{i}({x}_{0:i}) + \sum^t_{s=i} \Delta(s)$ for $i = t,\ldots,0$.\\
\\
\noindent
\emph{Base case} ($i = t$):
$$
\begin{array}{rl}
\widetilde{V}_{t}({x}_{t}) =& \displaystyle\max_{{a}_{t} \in \set{A}({x}_{t})}\mathbb{H}[{Z}_{\tau({x}_{t}, {a}_{t})}|{Z}_{{x}_{t}}]\vspace{1mm}\\
=& \displaystyle\mathbb{H}[{Z}_{\tau(x_t,\widetilde{\pi}_t(x_t))}|{Z}_{{x}_{t}}]\vspace{1mm}\\
\leq & \displaystyle\mathbb{H}[{Z}_{\tau(x_t,\widetilde{\pi}_t({x}_{t}))}|{Z}_{{x}_{0:t}}] +\Delta(t)\vspace{1mm}\\
=& {V}^{\widetilde{\pi}}_{t}({x}_{0:t}) +\Delta(t) \ .
\end{array}
$$
The first equality follows from (\ref{eq:9.2}).
The inequality follows from Lemma~\ref{lem:9.8}.
The last equality is due to (\ref{eq:9.9a}).
So, the base case is true.\\
\\
\noindent
\emph{Inductive case}: Suppose that 
\begin{equation}
\widetilde{V}_{i+1}({x}_{i+1}) \leq {V}^{\widetilde{\pi}}_{i+1}({x}_{0:i+1}) + \sum^t_{s=i+1} \Delta(s)
\label{eq:app19.1}
\end{equation}
is true. We have to prove that 
$\widetilde{V}_{i}({x}_{i}) \leq {V}^{\widetilde{\pi}}_{i}({x}_{0:i}) + \sum^t_{s=i} \Delta(s)$ is true.

By Lemma~\ref{lem:9.8},
$$
\begin{array}{rl}
&\displaystyle\mathbb{H}[{Z}_{{x}_{i+1}}|{Z}_{{x}_{i}}]  \leq \mathbb{H}[{Z}_{{x}_{i+1}}|{Z}_{{x}_{0:i}}]+\Delta(i) \ \ \ \mbox{for any} \ {x}_{i+1}\vspace{1mm}\\
\Rightarrow &\displaystyle\mathbb{H}[{Z}_{{x}_{i+1}}|{Z}_{{x}_{i}}] + \widetilde{V}_{i+1}({x}_{i+1}) \leq \mathbb{H}[{Z}_{{x}_{i+1}}|{Z}_{{x}_{0:i}}] \ +\vspace{1mm}\\ 
& {V}^{\widetilde{\pi}}_{i+1}({x}_{0:i+1}) + \sum^t_{s=i} \Delta(s) \ \mbox{by} \ (\mbox{\ref{eq:app19.1}}) \ \mbox{for any} \ {x}_{i+1}\vspace{1mm}\\
\Rightarrow &\displaystyle\mathbb{H}[{Z}_{\tau(x_i,\widetilde{\pi}_i(x_i))}|{Z}_{{x}_{i}}] + \widetilde{V}_{i+1}(\tau(x_i,\widetilde{\pi}_i(x_i)))\vspace{1mm}\\ 
& \leq \displaystyle\mathbb{H}[{Z}_{\tau(x_i,\widetilde{\pi}_i({x}_{i}))}|{Z}_{{x}_{0:i}}] +{V}^{\widetilde{\pi}}_{i+1}(
\ ({x}_{0:i}, \tau({x}_{i}, \widetilde{\pi}_i({x}_{i})))\ 
) \ + \vspace{1mm}\\
& \ \ \ \sum^t_{s=i} \Delta(s) \mbox{by} \ {x}_{i+1} \leftarrow \tau(x_i,\widetilde{\pi}_i(x_i))\vspace{1mm}\\
\Rightarrow &\displaystyle\widetilde{V}_{i}({x}_{i}) \leq {V}^{\widetilde{\pi}}_{i}({x}_{0:i}) + \sum^t_{s=i} \Delta(s) \ \mbox{by} \ (\mbox{\ref{eq:9.2}}) \ \mbox{and} \ (\mbox{\ref{eq:9.9a}}).
\end{array}
$$
Hence, the inductive case is true.
\end{document}